\pdfoutput=1
\documentclass{article}
\usepackage{arxiv}
\usepackage{times}
\usepackage{natbib}
\setcitestyle{authoryear,round,citesep={;},aysep={,},yysep={;}}
\usepackage{amsmath,amsfonts,bm}

\def\eqref#1{equation~\ref{#1}}

\def\1{\bm{1}}

\DeclareMathAlphabet{\mathsfit}{\encodingdefault}{\sfdefault}{m}{sl}
\SetMathAlphabet{\mathsfit}{bold}{\encodingdefault}{\sfdefault}{bx}{n}

\usepackage[T1]{fontenc}
\usepackage[utf8]{inputenc}
\usepackage{graphicx,booktabs,array,xcolor,capt-of,placeins,microtype,needspace}
\usepackage{overpic}
\usepackage{hyperref}
\hypersetup{hidelinks,hypertexnames=false}
\usepackage{url,textcomp}
\DeclareUnicodeCharacter{2212}{\ensuremath{-}}
\DeclareUnicodeCharacter{200B}{}
\DeclareUnicodeCharacter{03B8}{\ensuremath{\theta}}
\DeclareUnicodeCharacter{0394}{\ensuremath{\Delta}}
\graphicspath{{figures/}}
\title{Learning from Natural Images\\Untangles Number Sense}
\author{%
Yechan Cho$^{1,2,*}$, Hyeonsu Lee$^{1,*}$, Emma Deneuville$^{1,3}$, and Jaeson Jang$^{1,2}$\\[2pt]
\small $^{1}$Center for Brain Convergence Research, Brain Science Institute,\\
\small Korea Institute of Science and Technology, Seoul, Republic of Korea\\
\makebox[0pt][c]{\small $^{2}$Division of Bio-Medical Science \& Technology, University of Science and Technology, Daejeon, Republic of Korea}\\
\small $^{3}$Universit\'e C\^ote d'Azur, Institut NeuroMod, France\\
\small $^{*}$These authors contributed equally.}
\date{}
\renewcommand{\shorttitle}{Learning from Natural Images Untangles Number Sense}
\renewcommand{\headeright}{A Preprint}
\renewcommand{\undertitle}{A Preprint}
\hypersetup{pdftitle={Learning from Natural Images Untangles Number Sense},pdfauthor={Yechan Cho, Hyeonsu Lee, Emma Deneuville, Jaeson Jang}}
\begin{document}
\maketitle
\begin{abstract}
Numerical cognition is a simple yet fundamental ability that remains challenging for modern vision-language models. Meanwhile, number sense is already present in numerically naïve animals, suggesting that it can emerge without explicit numerical learning. Deep neural networks provide a related observation, with numerosity-selective units appearing after training on unrelated natural-image tasks and even before any training. However, these findings leave unresolved whether visual learning contributes beyond the spontaneous emergence of numerical selectivity. Human development provides a clue, as children become better at making judgments in which numerosity and proportion conflict. Here, we introduce a geometric framework to trace how representations of these two quantities change across learning. Although untrained and pretrained networks exhibited similar unit-level numerosity selectivity and basic comparison performance, natural-image pretraining increased the angle between numerosity and proportion subspaces while reducing their overlap, supporting more robust readout under conflict. Direct manipulation of these geometric properties further demonstrated their functional contributions to decoding accuracy and functional independence. Importantly, conventional supervised pretraining was not essential for quantity separation, which also emerged through self-supervised learning without category labels and synthetic-image training without natural-image exposure. Together, our findings suggest that structured visual learning can refine spontaneously emerging number sense into robust and functionally independent quantity representations without explicit numerical training.

\end{abstract}

\keywords{number sense \and numerical cognition \and representation learning \and representational geometry}

\paragraph{TL;DR} Natural-image learning enables robust number sense by separating numerosity and proportion representations.

\section{Introduction}

Numerical cognition supports estimation, comparison, and counting, and provides a foundation for more complex quantitative reasoning \citep{nieder2009representation}. Yet despite broad visual pretraining, modern vision-language models remain unreliable at object counting and related quantitative judgments \citep{paiss2023teaching,sengupta2026count}. In biological systems, however, sensitivity to numerical quantity emerges without explicit numerical instruction. Numerosity-selective neurons are observed in numerically naïve monkeys, and human infants can discriminate ratios between visual quantities \citep{viswanathan2013neuronal,mccrink2007ratio}. These observations raise a fundamental question of how robust quantity representations emerge without direct numerical training.

\begin{figure}[!t]
\centering
\includegraphics[width=0.78\linewidth]{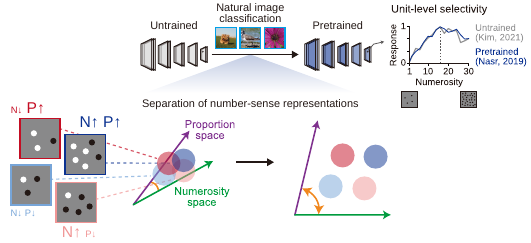}
\caption{Natural-image pretraining reorganizes number-sense representations at the population level. Although similar numerosity selectivity has been reported at the unit level in untrained and pretrained DNNs, we show that visual learning substantially reorganizes population geometry, separating numerosity and proportion representations for more independent readout.}
\label{fig:main-1}
\end{figure}

Artificial neural networks provide a controlled system for studying how learning experience contributes to the emergence and refinement of number sense. Previous work has shown that numerosity-selective units emerge after training on unrelated object-recognition tasks \citep{nasr2019number}, and even in randomly initialized networks that have never been trained \citep{kim2021visual}. Untrained networks can also support comparisons of other visual quantities, including proportion and difference \citep{lee2023comparison}. These findings establish that basic quantity sensitivity does not require numerical supervision. What remains unclear is whether ordinary visual learning simply preserves this numerical selectivity or reorganizes it to support more robust population-level decoding.

Human development suggests a more demanding test of this distinction. In proportional reasoning, absolute counts can interfere with proportion judgments when the two cues conflict, particularly for discrete quantities \citep{jeong2007development,boyer2008development}. Importantly, children become better able to resolve conflicting quantity cues over development, suggesting a developmental refinement of the representations supporting numerical judgments \citep{hamamouche2023winning}. Resolving such conflicting cues requires one quantity to be independently extracted despite variation in the other quantity, suggesting that the two quantities need to be represented along separable dimensions at the population level beyond single-unit selectivity.

Recent work provides a geometric framework for addressing this question. Rather than characterizing representations solely by single-unit tuning, population-level analyses quantify how responses to different input conditions are arranged in a multidimensional space. Measures such as the separation, overlap, distance, and relative orientation of condition-specific representations have been used to track how decodable structure emerges and how well it generalizes across conditions \citep{rigotti2013importance,bernardi2020geometry,cheng2025neural,styves2025variation,wakhloo2026neural}. Inspired by this perspective, we examine not only whether numerosity and proportion information is present at the level of individual units, but how their relative geometry changes with visual learning and whether this reorganization supports more robust decoding.

Here, we trace the representational geometry of numerosity and proportion across learning. Although both quantities are already represented at the unit level in untrained networks, visual learning without an explicit numerical objective separates their population representations by increasing subspace angles and reducing representational overlap, improving decoding when the cues conflict. We directly manipulated these geometric properties to validate their functional contributions to quantity readout, demonstrating that accurate decoding does not necessarily imply functional independence. Similar separation also emerged through label-free self-supervised learning and structured synthetic-image training. Our main contributions are as follows:

\begingroup
\setlength{\leftmargini}{0.5\leftmargini}
\begin{itemize}
\setlength{\itemsep}{0pt}
\item %
We introduce a representational geometry framework to show that natural-image pretraining refines number sense by separating numerosity and proportion representations.

\item %
We functionally validate this geometric account by directly manipulating representational geometry and observing the predicted changes in quantity readout.

\item %
We show that quantity separation emerges even without category labels or natural images, suggesting that structured visual learning alone can support the refinement of number sense.

\end{itemize}
\endgroup

\clearpage
\section{Experimental Setup}\label{sec:setup}

\subsection{Networks and visual experience}

\textbf{Architectures.} To test generality, we analyzed untrained and pretrained versions of twelve models: CNNs (AlexNet \citep{krizhevsky2012imagenet}, ResNet-50, ResNeXt-50, EfficientNet-B0, DenseNet121), vision transformers (ViT-B/16, Swin-T, DeiT III-L/16), a vision-language encoder (UniCL Swin-T), and other designs (gMLP-S, Vim-S, PoolFormer-S24). Three of them also had self-supervised variants (SEED, DINO, MoBY; \citealp{fang2021seed,caron2021emerging,xie2021self}). Training details are in Appendix~\ref{app:training}.

\textbf{Training states.} We compared ten training states: randomly initialized (Untrained); ImageNet-1K supervised classification (Pretrained); the same training with shuffled labels; a Developmental Visual Diet (DVD; \citealp{lu2026adopting}); self-supervised SimCLR \citep{chen2020simple} and RotNet \citep{gidaris2018unsupervised}; and SimCLR on four synthetic image sets of increasing structure: noise, gratings, statistical images \citep{baradad2021learning}, and procedural Shaders \citep{baradad2022procedural}.

\subsection{Quantity stimuli and representational geometry}

\textbf{Stimuli.} Each image contained $N_W$ white and $N_B$ black dots (Figure~\ref{fig:supp-1}). Numerosity is the total count, $N=N_W+N_B$ (1 to 30), and proportion is the white fraction, $P=N_W/N$. 18,000 images covered $(N_W,N_B)$ combinations under three regimes controlling dot size, total area, or extent.

\textbf{Quantity subspaces.} We analyzed frozen activations $\mathbf x$: conv5 responses for AlexNet and the final pooled or CLS features for the other architectures. For each quantity, we averaged the activations over the other quantity and applied PCA. The leading components span the numerosity subspace $\mathbf V_N$ and the proportion subspace $\mathbf V_P$. We used one $N$ and two $P$ components for AlexNet and two components each for the other architectures, fixed across the states of each model so that geometry is always compared at the same dimensionality (Figure~\ref{fig:supp-2}).

\textbf{Two measures of separation.} The \emph{principal angle} describes how differently the two subspaces are oriented, $\theta=\cos^{-1}\sigma_{\max}(\mathbf V_N^{\top}\mathbf V_P)$, from $0^{\circ}$ (aligned) to $90^{\circ}$ (orthogonal). \emph{Representational overlap} describes how much the responses to different $(N,P)$ conditions overlap. We fitted a Gaussian to each condition in the joint $N$/$P$ subspace and averaged the Bhattacharyya coefficient over all pairs of conditions on held-out data, from 0 (separable) to 1 (identical) (Appendix~\ref{app:angle},~\ref{app:geometry}).

\subsection{Geometric manipulation and cross-quantity ablation}

\textbf{Manipulating geometry.} To test whether this geometry matters for readout, we edited the activations of twelve pretrained networks (AlexNet and the eleven additional architectures), leaving images and weights unchanged. We treated this as an inverse problem: given a target geometry, we sought a transformation of the activations that reaches it while changing only one property. All edits were confined to the joint $N$/$P$ subspace, with orthonormal basis $\mathbf Q$. To change the principal angle, we fitted a smooth deformation (embedded deformation with as-rigid-as-possible regularization; \citealp{sumner2007embedded,sorkine2007arap}) that shifted the angle by $-15^{\circ}$ to $+15^{\circ}$ in $5^{\circ}$ steps and left overlap nearly unchanged. To change overlap, we rescaled the spread of the responses around each condition mean $\boldsymbol\mu_c$ by a factor $a\in\{0.25,0.5,1,2,4\}$,
\begin{equation}
\mathbf x^{(a)}=\mathbf x+(a-1)\,\mathbf Q\mathbf Q^{\top}(\mathbf x-\boldsymbol\mu_c),
\end{equation}
which leaves the condition means, and hence the angle, unchanged (Figure~\ref{fig:main-4}a; Appendix~\ref{app:geometry_interventions}).

\textbf{Two functional readouts.} After each edit, we measured decoding accuracy by refitting linear decoders of $N$ and $P$, and cross-quantity ablation damage, which asks how much the readout of one quantity relies on the subspace of the other (Figures~\ref{fig:supp-3} and~\ref{fig:supp-4}). For a target quantity $q$, we removed the variation along the other subspace $\mathbf V_{\bar q}$ while preserving the mean activation $\boldsymbol\mu$,
\begin{equation}
\mathbf x_{\mathrm{ablated}}=\mathbf x-\mathbf V_{\bar q}\mathbf V_{\bar q}^{\top}(\mathbf x-\boldsymbol\mu),
\end{equation}
and defined damage as the resulting increase in the error of the decoder, which was held fixed. Removing a random subspace of the same rank served as a control (Appendix~\ref{app:ablation}). For the AlexNet training states, we also related geometry to joint $(N,P)$ decoding, comparison under $N$/$P$ conflict, and cross-condition generalization (Appendix~\ref{app:function}).

\section{Natural-image pretraining refines number sense under conflict}

We first asked whether the standard unit-level criterion of number sense distinguishes a trained network from an untrained one. Following \citet{nasr2019number} and \citet{kim2021visual}, we presented single-color dot arrays spanning numerosities from 1 to 30 to an ImageNet-pretrained AlexNet and to the same architecture with randomly initialized weights. Figure 2a shows that both untrained and pretrained networks contained conv5 units selectively tuned to numerosity across all three stimulus sets controlling for dot size, total area, and convex hull, consistent with previous reports \citep{nasr2019number,kim2021visual}. Although pretraining sharpened the tuning of number-selective units (selectivity index, untrained vs. pretrained: 0.569 vs. 0.948) while slightly reducing their fraction (7.7\% vs. 5.7\%), it remained unclear whether these unit-level changes translated into improved population-level decoding.

We therefore assessed population-level numerosity readout using a comparison task following \citet{nasr2019number} and \citet{kim2021visual}. A linear classifier trained on conv5 activity predicted which of two dot arrays contained the larger number of dots. 
Figure 2b shows that, when tested with white dots on a black background as in \citet{kim2021visual}, both untrained and pretrained networks performed well above chance (50\%), with only a small advantage after pretraining (84.5\% vs. 85.4\%). To examine whether this finding generalized to more diverse visual conditions, we extended the task to eight color and background conditions, including mixed black-and-white, grayscale, red/blue, and multicolored dots (Figure S5). Pretrained networks showed a modest advantage of 0.9–6.4 percentage points across these conditions, but untrained networks still maintained high accuracy (79.5–84.7\%), even with mixed-color stimuli (Figure 2b, right). Thus, although pretraining modestly improves basic numerosity discrimination, both networks can reliably extract numerosity from diverse visual inputs, motivating a more demanding test to reveal how number sense develops beyond basic numerical sensitivity.

\begin{figure}[!b]
\centering
\includegraphics[width=1.00\linewidth]{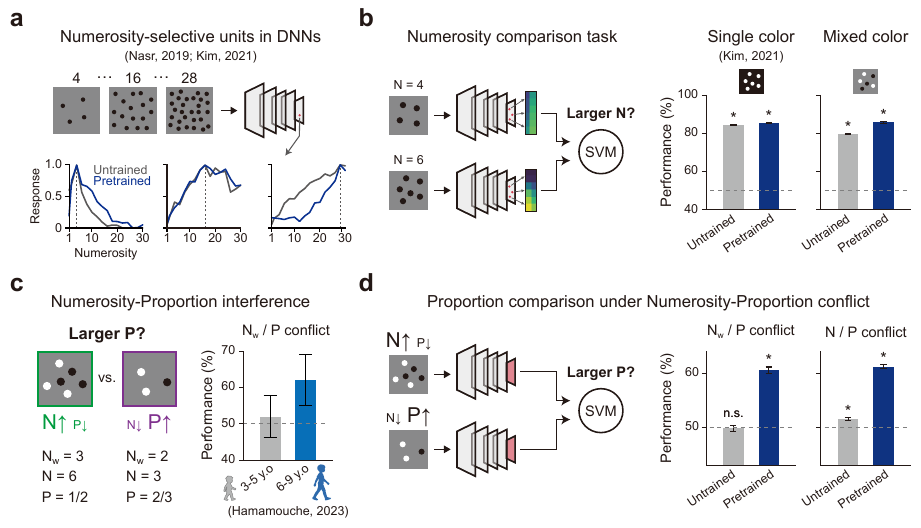}
\caption{Basic numerosity sensitivity is similar across untrained and pretrained networks, whereas quantity conflict reveals a functional difference. (a) Both untrained and ImageNet-pretrained AlexNet contain conv5 units selectively tuned to numerosity across single-color dot arrays. (b) A linear SVM comparing which of two arrays has the larger numerosity achieves similarly high performance in both network states. (c) Human developmental studies show that proportion judgments become more robust with age when absolute count and proportion provide conflicting cues. (d) We adapted this conflict paradigm to two-color dot arrays, asking which array has the larger proportion when numerosity and proportion favor opposite choices. The untrained network performs near chance, whereas the pretrained network remains above chance.}
\label{fig:main-2}
\end{figure}

A clue came from human developmental research on proportional reasoning, in which absolute counts can interfere with judgments of proportion. \citet{hamamouche2023winning} examined how children aged three to nine compared the relative proportions of different colors in two images. Figure 2c illustrates their developmental finding: when proportion and absolute count conflicted, younger children performed near chance, whereas older children were better able to resolve the conflict. These findings suggest that numerical cognition becomes more robust over development, marking a more mature form of numerical cognition.

Inspired by this paradigm, we compared untrained and pretrained networks on a more demanding proportion comparison task involving conflicting quantity cues. Using pairs of two-color dot arrays, we trained a linear readout on conv5 activity to identify the array with the higher proportion of white dots ($P$). Following \citet{hamamouche2023winning}, we initially defined congruency based on white-dot count ($N_W$) and proportion ($P$): trials were congruent when the array with a higher $P$ also had a higher $N_W$, and incongruent when it had a lower $N_W$. Hereafter, we refer to incongruent trials as "conflict" trials, specifying $N_W/P$ or $N/P$ according to whether the conflict involves white-dot count or total numerosity. Under $N_W/P$ conflict, the untrained network performed near chance (49.9\%), whereas the pretrained network reached 60.6\% accuracy (Figure 2d; Figure S6).

To examine a more general form of quantity conflict, we extended this analysis to total numerosity ($N$) and proportion ($P$), because $N_W/P$ is included in $N/P$ conflict (Figure S7). Figure 2d shows a similar pretraining advantage under this broader definition, which also persisted under additional controls for low-level visual features (Figure S8). Given its broader scope, we focused on $N/P$ conflict in subsequent analyses of population-level representations. Hereafter, all comparison tasks use the hard setting, in which image pairs differ by only one discrete step in both numerosity ($|\Delta N|=1$) and proportion ($|\Delta P|=1/6$), as in Figure 2d.

\section{Pretraining separates the numerosity and proportion subspaces}

To characterize how visual pretraining reorganizes quantity representations, we compared numerosity and proportion geometry across 12 models, ranging from AlexNet and ResNet, widely used in previous number-sense studies, to ten modern architectures. Figure 3a shows the two-color stimulus space, spanning numerosity values from $N$ = 1 to 30 and proportion values from $P$ = 0 to 1 in steps of 1/6. For each network, feature responses were averaged across variation in the other quantity and then subjected to PCA. The retained components captured most of the condition-dependent variance. Within each model, the subspace dimensionality was kept the same across training states, so that changes in geometry reflect the representation rather than the dimensionality (Figure S2).

If the improved conflict performance after pretraining reflects a more separable population representation, numerosity- and proportion-related variation should become more distinct in representational space. Figure 3b illustrates two complementary aspects of representational separation: the relative orientation of the N and P subspaces and the overlap among individual ($N$, $P$) response distributions. We quantified these using the principal angle and Gaussian distributional overlap, respectively.

Figure 3c shows that natural-image pretraining consistently increased the principal angle between the $N$ and $P$ subspaces while reducing representational overlap across all 12 models tested. This pattern held across diverse architectures, including CNNs, Vision Transformers, a vision-language model encoder, and alternative architectures. Importantly, the same geometric separation was observed when the three stimulus sets were analyzed separately rather than pooled (Figure S8), indicating that the findings were not driven by pooling stimuli with different low-level visual statistics.

Together, these results demonstrate that visual pretraining consistently reorganizes quantity representations across diverse architectures and stimulus conditions, even without an explicit numerical learning objective. This geometric reorganization provides a potential population-level explanation for the improved conflict performance in Figure 2, which we further investigate through direct manipulation of both geometric properties.

\begin{figure}[!ht]
\centering
\includegraphics[width=0.89\linewidth]{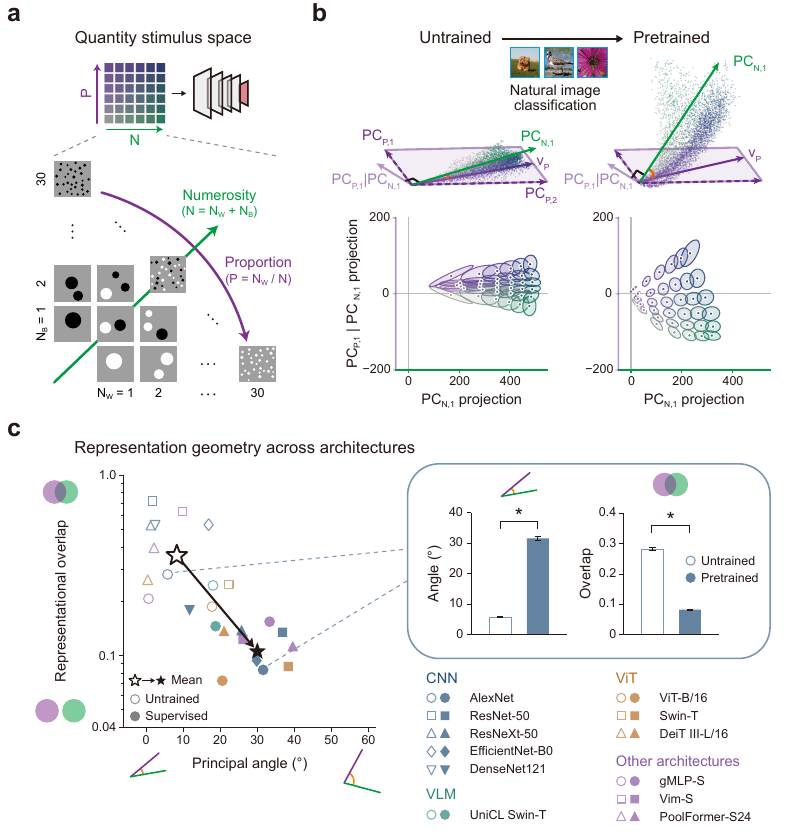}
\caption{Natural-image pretraining separates numerosity and proportion representations at the population level. (a) Stimulus design spanning numerosity and proportion across two-color dot arrays. (b) PCA of AlexNet conv5 responses reveals a one-dimensional numerosity subspace and a two-dimensional proportion subspace, which become more separated after pretraining. (c) Across diverse network architectures, this reorganization is quantified by increased principal angles and reduced representational overlap, suggesting that visual learning without an explicit numerical objective produces more separable quantity representations.}
\label{fig:main-3}
\end{figure}

\section{Manipulating representational geometry reveals its functional role in quantity readout}

To directly test whether representational geometry functionally contributes to quantity readout, we artificially manipulated two geometric properties, principal angle and representational overlap, within individual pretrained networks. Figure 4a illustrates our approach, in which we treated the manipulation as an inverse problem, applying smooth transformations to the original population activations to achieve desired changes in representational geometry. Specifically, we increased or decreased principal angle while approximately preserving representational overlap, and vice versa, allowing us to examine their functional contributions independently (see Appendix A.7 for methods).

\begin{figure}[!tbp]
\centering
\includegraphics[width=1.00\linewidth]{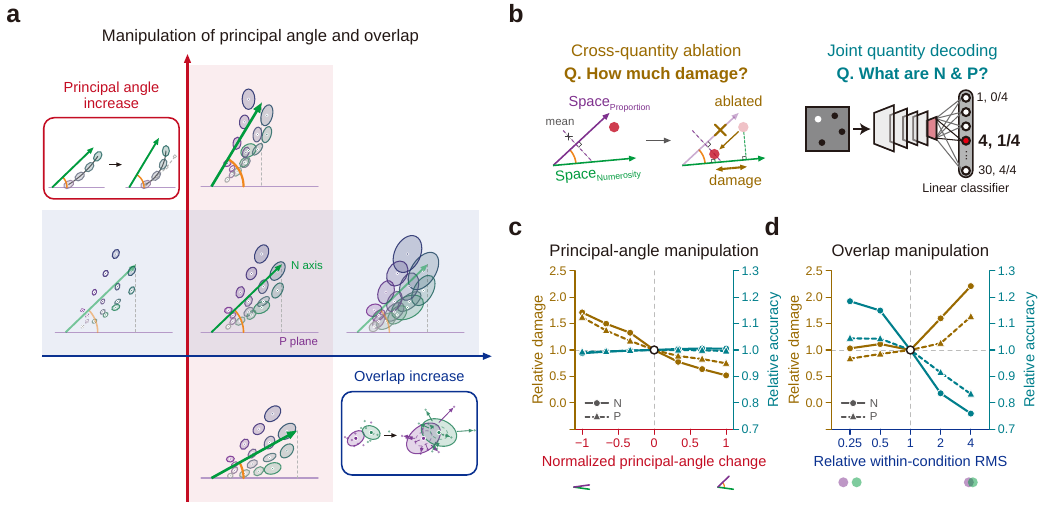}
\caption{Functional validation of representational geometry through direct manipulation. (a) Population activations of pretrained networks were transformed to independently manipulate the principal angle between numerosity ($N$) and proportion ($P$) subspaces or their representational overlap, while approximately preserving the other property. (b) Functional evaluation using joint quantity decoding accuracy and cross-quantity ablation damage. (c) Increasing principal angle reduced ablation damage, with little effect on decoding accuracy. (d) Increasing representational overlap reduced decoding accuracy and increased ablation damage. Values in (c,d) are normalized to the original representations and averaged across 12 architectures.}
\label{fig:main-4}
\end{figure}

Because principal angle and representational overlap capture different aspects of quantity representations, we examined their effects on two complementary functional measures: decoding accuracy and cross-quantity ablation damage (Figure 4b). We measured decoding accuracy using linear classifiers to read out $N$ and $P$ from the manipulated activations. To assess functional independence, we performed cross-quantity ablation, removing one quantity's subspace and measuring the resulting damage to decoding the other. We used mean-preserving ablation to avoid activation mean shifts, with matched-rank random-subspace ablation as a control (Appendix A.8; Figure S3).

Figure 4c shows that increasing principal angle reduced cross-quantity ablation damage, whereas decreasing the angle increased damage, with little effect on decoding accuracy. In contrast, Figure 4d shows that increasing representational overlap reduced decoding accuracy while also increasing ablation damage (see Figure S9 for individual architectures). Thus, decoding accuracy was primarily affected by representational overlap, whereas functional independence could be independently modulated by changing principal angle without altering decoding accuracy. Together, these intervention-based findings demonstrate that accurate decoding does not necessarily imply functionally independent quantity representations, highlighting the importance of examining subspace organization beyond decoding performance alone.

\section{Quantity separation extends beyond conventional supervised pretraining}

Having found that supervised natural-image pretraining reorganizes numerosity and proportion representations, we next asked whether quantity separation could emerge beyond conventional supervised pretraining. We therefore examined self-supervised learning (SSL) without category labels and training on synthetic images without natural-image exposure.

We first examined networks trained with self-supervised learning (SSL) on natural images (AlexNet: SimCLR \citep{chen2020simple}, RotNet \citep{gidaris2018unsupervised}; EfficientNet-B0: SEED \citep{fang2021seed}; ViT-B/16: DINO \citep{caron2021emerging}; Swin-T: MoBY \citep{xie2021self}). Figure 5 shows that these SSL models consistently exhibited greater representational separation than their untrained counterparts, with larger principal angles between the N and P subspaces and lower representational overlap. The consistency of this pattern across diverse SSL models indicates that explicit category labels are not a prerequisite for quantity separation.

\begin{figure}[!tbp]
\centering
\includegraphics[width=1.00\linewidth]{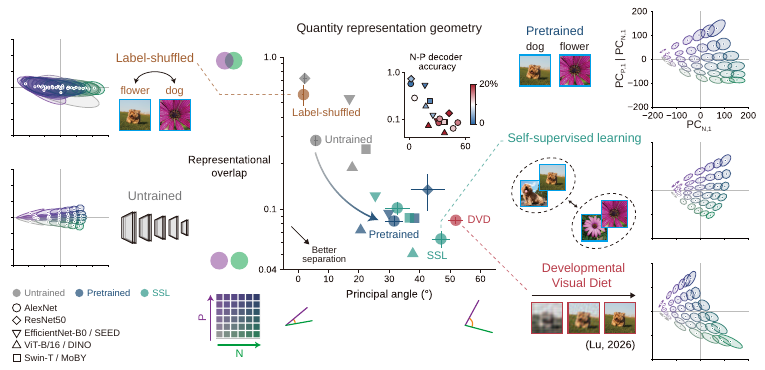}
\caption{Quantity separation emerges without category labels but not under arbitrary supervision. Self-supervised learning produces a similar reorganization to supervised pretraining, whereas shuffled-label training reduces separation below the untrained baseline. Developmentally restricted visual input also supports strong quantity separation.}
\label{fig:main-5}
\end{figure}

At the opposite extreme, we asked whether incorrect learning could impair the development of number sense. Figure 5 also shows the opposite effect of shuffled-label training: the principal angle decreased and representational overlap increased beyond the untrained baseline. This indicates that maladaptive learning can disrupt the emergence of separable quantity representations even when natural-image exposure is preserved (see Figure~\ref{fig:supp-10} for details).

\begin{figure}[!b]
\centering
\includegraphics[width=1.00\linewidth]{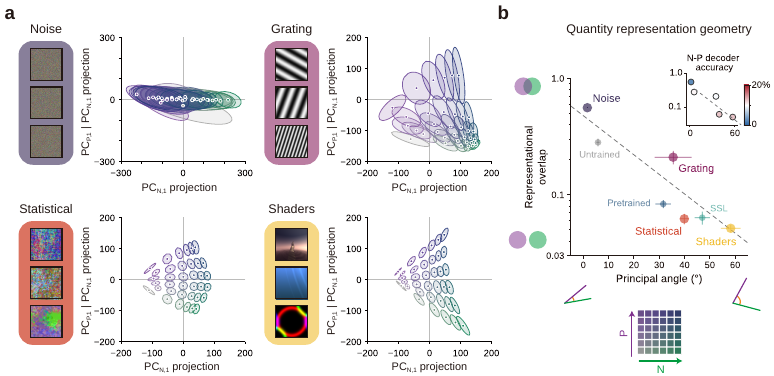}
\caption{Structured synthetic visual experience is sufficient to separate quantity representations. (a) Self-supervised learning on synthetic images ranging from noise to shader-based patterns produces progressively more organized quantity representations. (b) Representational separation increases with richer visual structure, showing that natural images are not uniquely required for quantity reorganization.}
\label{fig:main-6}
\end{figure}

We next asked whether quantity separation could still emerge under visual conditions that better approximate early development, when information such as contrast and acuity is substantially limited. We therefore examined a Developmental Visual Diet (DVD) model, in which visual acuity, contrast sensitivity, and chromatic sensitivity gradually mature over training \citep{lu2026adopting}. Figure 5 further shows that the DVD-trained network retained this reorganization despite early visual restrictions, with larger subspace angles and lower condition-wise overlap than the untrained state. Thus, these results suggest that the reorganization of quantity representations is compatible with a developmental pathway that more closely approximates the constraints of biological visual experience.

We next asked whether natural images themselves were necessary for quantity separation. Figure 6a shows the synthetic training distributions, ranging from noise and gratings to statistical images containing diverse frequency and color information and Shader images containing varied geometric patterns. Figure 6b shows that quantity separation increased across increasingly structured visual inputs, following the order Noise < Grating < Statistical < Shader (Figure~\ref{fig:supp-11}). Notably, Shader-based training produced even stronger separation than self-supervised learning on natural images. Thus, natural images are not uniquely required for quantity reorganization; rather, sufficiently rich and structured visual experience appears to be enough to disentangle numerosity and proportion representations.

Together, these results show that quantity separation extends beyond conventional supervised pretraining and is not restricted to natural images. It emerged without explicit category labels, persisted under restricted early visual input, and increased with the structure of synthetic training images, whereas shuffled-label learning reduced separation. These findings suggest that structured visual learning can refine quantity representations without explicit numerical instruction or category supervision (Figure~\ref{fig:supp-12}).

\section{Additional functional validation across training states}

Finally, to complement our direct geometric manipulations, we examined whether the geometry–function relationships held across different architectures and training conditions. We focused on two functional measures: joint $N$–$P$ decoding and numerosity comparison under conflicting quantity cues. Across diverse architectures, lower condition-wise overlap was associated with better joint decoding (Fig. 7a) and higher conflict-comparison accuracy (Fig. 7b). We observed similar relationships across the AlexNet training conditions examined in Section 6 (Figs. 7c,d; Tables 1,3). Together, these results show that the relationship between representational geometry and functional performance extends across both architectural differences and training conditions, providing further support for the role of representational separation in resolving competing quantity information.

\begin{figure}[!t]
\centering
\includegraphics[width=1.00\linewidth]{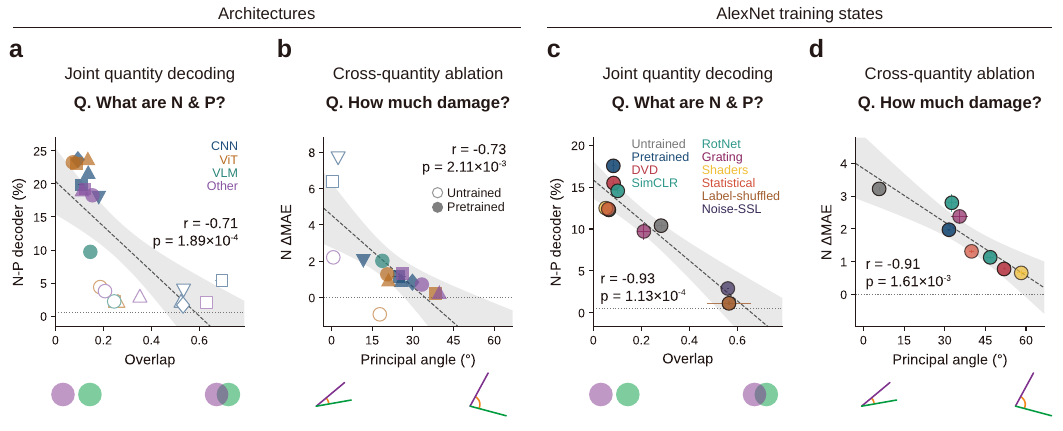}
\caption{Functional validation of quantity representation geometry. (a,b) Lower representational overlap is associated with better joint quantity decoding and higher accuracy under N--P conflict. (c,d) Larger principal angles are associated with better generalization to unseen quantity conditions and less damage following cross-quantity ablation.}
\label{fig:main-7}
\end{figure}

However, representational geometry did not fully predict performance across all training states (Figures S6 and S13). Under $N_W/P$ conflict, the Noise-trained network achieved higher incongruent accuracy than the pretrained network (72.5\% vs. 60.6\%) despite substantially greater overlap (0.561 vs. 0.083). Nevertheless, its congruent accuracy (52.3\% vs. 78.9\%) and overall accuracy (56.4\% vs. 75.2\%) were considerably lower. Its baseline decoding errors were also no better than those of a constant-median predictor (Figure S4), indicating that its incongruent advantage did not reflect generally better quantity readout. Although the source of this exception remains unclear, these findings suggest that representational geometry contributes to quantity readout but does not alone determine performance across different training conditions.

\section{Conclusion}

Number sense has traditionally been studied through numerosity-selective responses in biological and artificial systems, yet such selectivity alone does not explain whether quantity information can be used robustly when multiple cues interact. Recent work on multidimensional population geometry has similarly emphasized that computation depends not only on what information is encoded, but also on how variables are organized in representational space \citep{rigotti2013importance,bernardi2020geometry,cheng2025neural,styves2025variation,wakhloo2026neural}. Our results extend this perspective to number sense, showing that visual learning without an explicit numerical objective can transform initially available quantity sensitivity into more separable representations that support robust and independent readout.

A broader implication is that this reorganization is not tied to a particular training objective or to natural photographs themselves. Numerical variables that are never specified by the training objective can nevertheless become more separable as a network learns regularities in visual input. Importantly, this separation does not appear to depend on a single form of supervision or on natural photographs as a unique source of experience. Together, these findings suggest that structured learning may reorganize incidental information into representations better suited for downstream use.

More generally, these findings suggest that learning can substantially reorganize representations of variables that are not directly related to the training objective. This provides a possible geometric perspective on transfer learning, in which source-task experience can improve downstream capabilities by shaping latent representations before the downstream variable is ever explicitly trained \citep{yosinski2014transferable,neyshabur2020transferred,ericsson2021transfer}. Rather than evaluating training data and objectives only by performance on the source task, it may therefore be useful to consider how they organize representations that can support generalization to unseen tasks and domains. In this view, the emergence of separable latent geometry may be one factor underlying the broader benefits of pretraining and transfer.

These results also motivate extensions to both biological and multimodal systems. In biological vision, developmental recordings could test how ordinary visual experience shapes the population geometry of quantity representations over time. In VLMs, future work could trace how quantity representations evolve from visual encoders through multimodal and language layers, identifying where numerical information becomes entangled, inaccessible, or lost during processing. Such analyses could help explain persistent counting failures and reveal representational bottlenecks that may be targeted by future training strategies. More broadly, number sense provides a compact setting for studying how experience transforms latent information into representations that can be flexibly and independently used.

\label{end:main-text}
\clearpage
\subsection*{AI use statement}

Generative AI tools were used to provide feedback on experimental methodology and interpretation of results, and to assist with manuscript organization, editing, and the presentation of analyses. They were also used to aid and polish writing and for retrieval and discovery of relevant literature and related work. All AI-assisted suggestions were reviewed by the authors, and all reported analyses and results were independently verified. The authors take responsibility for the final content of this work.

\subsection*{Reproducibility statement}

Detailed descriptions of the network architectures, training procedures, stimulus generation, representational analyses, functional readouts, statistical tests, and replication settings are provided in Section 2, Experimental Setup, and the Appendix. The number of independent runs and evaluation procedures are reported for each condition to facilitate reproduction of the results. The data generated for this study and the code used to generate the stimuli and analyze the results are available upon request. 

\subsection*{Acknowledgments}

This work was supported by KIST Intramural grants (26E0124, 26E0550, 26Z9002, 2E33701, 2E33721) and the National Research Foundation of Korea (RS-2024-00460958).

\phantomsection
\label{start:references}
\bibliography{references}
\bibliographystyle{plainnat}
\clearpage
\appendix
\setcounter{equation}{0}
\renewcommand{\theequation}{A\arabic{equation}}
\setcounter{figure}{0}
\renewcommand{\thefigure}{S\arabic{figure}}
\section{Detailed methods}
\label{app:methods}

\subsection{Networks and visual training}
\label{app:training}
We used AlexNet with matched He-normal initializations \citep{he2015delving}:
20 networks each for Untrained, Pretrained, Label-shuffled, and DVD, and five
each for SimCLR, RotNet, Grating, Shaders, Statistical, and Noise. Training
datasets contained 1,281,167 images, matching ImageNet-1K \citep{deng2009imagenet},
and supplied no numerical supervision. Geometry and task analyses used matched
checkpoints: epoch 0 for Untrained,  90 for supervised, synthetic, and natural-image 
SimCLR conditions, and 30 for RotNet.

Supervised training used SGD for 90 epochs (batch size 512, initial learning
rate 0.02, momentum 0.9, weight decay $5\times10^{-4}$). Label-shuffled training
used a fixed random permutation of labels across training images, preserving the empirical class-frequency distribution. DVD
\citep{lu2026adopting} retained this protocol while scheduling acuity, color,
and contrast over 0--300 simulated months.

SimCLR \citep{chen2020simple} used a $4096\rightarrow4096\rightarrow128$
projection head and LARS \citep{you2017large} (batch size 256, learning rate 0.3,
momentum 0.9, weight decay $10^{-6}$, temperature 0.5). A 10-epoch linear warmup
preceded cosine decay over 90 epochs.
Two $224\times224$ views used random resized crops, color jitter,
grayscale, Gaussian blur, and horizontal flips.
Synthetic datasets comprised sinusoidal Gratings, Spectrum + Color +
WMM images (Statistical; \citealp{baradad2021learning}), procedural Shaders
\citep{baradad2022procedural}, or fixed independent Gaussian pixel/channel
samples converted to RGB (Noise).

RotNet \citep{gidaris2018unsupervised} predicted $0^\circ$, $90^\circ$,
$180^\circ$, and $270^\circ$ rotations with a linear $4096\rightarrow4$ head,
averaging cross-entropy across views. Training used source batches of 192,
SGD momentum 0.9, weight decay $5\times10^{-4}$, and learning rates 0.01,
0.001, and 0.0001 over successive ten-epoch blocks. Quantity analyses used
frozen backbone features.

\subsection{Stimuli and frozen population responses}
\label{app:stimuli}
The common evaluation set contained 18,000 white/black-dot images covering
495 integer $(W,B)$ conditions with $1\leq W+B\leq30$, where $N=W+B$ and
$P=W/(W+B)$. Three stimulus regimes controlled dot size, total dot area, or
spatial extent. The same images were presented to every network. AlexNet
received $227\times227$ three-channel images with ImageNet input
normalization; the other architectures received them through their standard
input preprocessing.

Analyses used frozen responses in raw activation units, with $D$ the feature
dimension of each network. For AlexNet, we used all $D=43{,}264$ flattened
post-ReLU conv5 features. For the other architectures, we used the final
feature vector preceding the classification head: the globally pooled feature
for ResNet-50 ($D=2{,}048$), ResNeXt-50, EfficientNet-B0, DenseNet121,
Swin-T, gMLP-S, and PoolFormer-S24; the CLS token after the final
normalization for ViT-B/16, DeiT III-L/16, and Vim-S; and the pooled
vision-backbone output before the image--text projection for UniCL Swin-T.
Proportion used seven bins:
\begin{equation}
b(P)=\sum_{k=0}^{5}\mathbb I\!\left[P\geq\frac{2k+1}{12}\right],
\qquad b(P)\in\{0,\ldots,6\}.
\label{eq:pbin}
\end{equation}
Centers were $0,1/6,\ldots,1$, with midpoint boundaries assigned to the upper bin.

\subsection{Quantity subspaces and principal-angle search}
\label{app:angle}
Responses were averaged within each of the 30 N levels or seven P bins,
pooling over the other quantity. After centering across levels, separate PCA
of the level means gave orthonormal bases
$V_N\in\mathbb R^{D\times d_N}$ and $V_P\in\mathbb R^{D\times d_P}$ of the N
and P subspaces. For AlexNet, we retained the first N component and the first
two P components ($d_N=1$, $d_P=2$; $V_N=\mathbf u_N$) in every training
state, which captured more than 90\% of the condition-dependent variance in most
states (Figure~\ref{fig:supp-2}b,c). For the other architectures, we retained the first two N components and the
first two P components ($d_N=d_P=2$), which captured more than 90\% of the
condition-dependent variance in most states (Figure~\ref{fig:supp-2}d,e).
The dimensions were fixed across the states of each model, so that its
geometry was always compared at the same dimensionality.

Subspace orientation was the smallest principal angle between the two
subspaces, obtained from the largest singular value of $V_N^\top V_P$:
\begin{align}
\cos\theta
&=\max_{\|\mathbf a\|_2=\|\mathbf b\|_2=1}
 \left|\mathbf a^\top V_N^\top V_P\mathbf b\right|
 =\sigma_{\max}\!\left(V_N^\top V_P\right),\qquad
\theta=\arccos\sigma_{\max}\!\left(V_N^\top V_P\right),\\
\mathbf v_N&=V_N\mathbf a_\star,\qquad
\mathbf v_P=V_P\mathbf b_\star,
\label{eq:angle-search}
\end{align}
where $\mathbf a_\star$ and $\mathbf b_\star$ are the corresponding left and
right singular vectors and $(\mathbf v_N,\mathbf v_P)$ is the closest pair of
directions in the two subspaces. For AlexNet ($d_N=1$), this reduces to
$\cos\theta=\|V_P^\top\mathbf u_N\|_2$ with $\mathbf v_N=\mathbf u_N$ and
$\mathbf v_P=V_PV_P^\top\mathbf u_N/\|V_P^\top\mathbf u_N\|_2$.
Network-level principal angles used the full evaluation set.

\subsection{Held-out Gaussian overlap, distance, and volume}
\label{app:geometry}
Overlap was evaluated on held-out responses using three-fold cross-validation.
In each fold, $V_N$ and $V_P$ were fitted on the training responses, and the
held-out responses were projected into their joint span, of dimension
$d=d_N+d_P$. The projected responses were fitted per cell
$c$ and fold $f$ with Gaussian mean
$\boldsymbol\mu_{cf}$ and Ledoit--Wolf covariance $\Sigma_{cf}$
\citep{ledoit2004well}. For $K=495$ cells, NP overlap was
\begin{equation}
\mathrm{BC}_{NP}=\frac{1}{3}\sum_{f=1}^{3}
 \frac{2}{K(K-1)}\sum_{c<d}\mathrm{BC}
 \bigl(\mathcal N(\boldsymbol\mu_{cf},\Sigma_{cf}),
       \mathcal N(\boldsymbol\mu_{df},\Sigma_{df})\bigr),
\end{equation}
where, for two Gaussians $i$ and $j$ with $\Sigma=(\Sigma_i+\Sigma_j)/2$,
\begin{equation}
\mathrm{BC}_{ij}=\exp\!\left[-\frac18(\boldsymbol\mu_i-\boldsymbol\mu_j)^{\top}
\Sigma^{-1}(\boldsymbol\mu_i-\boldsymbol\mu_j)\right]
\sqrt{\frac{\sqrt{|\Sigma_i||\Sigma_j|}}{|\Sigma|}}.
\label{eq:bc}
\end{equation}

For N or P overlap, cells were pooled by N level or P bin with equal cell
weights. Suppressing the fold index, the class moments were
\begin{align}
\boldsymbol\mu_g&=\frac{1}{K_g}\sum_{c\in g}\boldsymbol\mu_c,\\
\Sigma_g&=\frac{1}{K_g}\sum_{c\in g}
 \left[\Sigma_c+(\boldsymbol\mu_c-\boldsymbol\mu_g)
                    (\boldsymbol\mu_c-\boldsymbol\mu_g)^\top\right].
\end{align}
BC was averaged over all class pairs and folds. Distance and 95\% ellipsoid
volume, reported for the AlexNet training states ($d=3$;
Table~\ref{tab:geometry-appendix}), used the same held-out cell statistics:
\begin{align}
D&=\frac{1}{3}\sum_f\frac{2}{K(K-1)}
       \sum_{c<d}\|\boldsymbol\mu_{cf}-\boldsymbol\mu_{df}\|_2,\\
V_{95}&=\frac{1}{3}\sum_f\frac{1}{K}\sum_c
       \frac{4\pi}{3}\left[\chi^2_3(0.95)\right]^{3/2}\sqrt{\det\Sigma_{cf}}.
\end{align}
BC is dimensionless; distance and volume retain raw activation units and
cubic units, respectively.

\subsection{Two-dimensional representation displays}
Bases fitted on folds B and C displayed fold A responses. We orthonormalized
either the N-PC1/P-PC1 directions or the closest principal-vector pair
$(\mathbf u_N,\mathbf v_P)$. Plots used six N groups of width five and the
seven P bins in Eq.~\ref{eq:pbin}; ellipses indicate one-standard-deviation
covariance contours. A network-specific activation scale normalized display
coordinates only.

\subsection{Functional readouts and comparison tasks}
\label{app:function}
\paragraph{Single-unit and comparison baseline.}
Following \citet{kim2021visual}, single-color dot arrays and two-way ANOVA
identified units with a numerosity effect at $p<0.01$ and stimulus-regime
and interaction effects at $p>0.01$ (uncorrected). A linear SVM compared
numerosity using 256 selected units; unit-draw results were averaged within
networks.

\paragraph{Direct N, P, and NP decoding.}
Affine softmax readouts without hidden layers decoded 30 N classes, seven
P bins, or 195 observed NP classes. A shared stratified 80/10/10\% split
defined training, validation, and test sets. AdamW
\citep{loshchilov2019decoupled} used learning rate and weight decay $10^{-5}$,
batch size 128, at most 200 epochs, and validation-loss early stopping with
patience 15. Performance was exact-class test accuracy.

\paragraph{Standard N/P pairwise comparison.}
Separate LinearSVC readouts predicted larger N or P from concatenated raw
responses $[\mathbf x_a;\mathbf x_b]$ using identical pairs and ordering.
Disjoint image pools supplied 800 training and 800 test pairs, including
400 N/P-congruent and 400 N/P-incongruent test pairs; congruency was
$\operatorname{sign}(N_a-N_b)=\operatorname{sign}(P_a-P_b)$.
Settings were $C=1$, hinge loss, L2 penalty, and at most 20,000 iterations,
with no feature scaling or PCA. Restricted comparisons trained and tested 
separate SVMs on pairs with $|N_a-N_b|=1$ for N comparison or 
$0.045\leq|P_a-P_b|\leq0.055$ for P comparison, again with 800 training 
and 800 test pairs.

\paragraph{Hard W/P comparison.}
A separate SVM predicted larger P for pairs one P bin apart, with congruency
defined by absolute white-dot count:
\begin{equation}
\operatorname{congruent}_{WP}(a,b)
\iff \operatorname{sign}(W_a-W_b)=\operatorname{sign}(P_a-P_b).
\end{equation}
The test set contained 640 congruent and 160 incongruent pairs, giving
$A_{\rm total}=0.8A_{\rm congruent}+0.2A_{\rm incongruent}$.

\paragraph{Bidirectional range generalization.}
N comparisons transferred between $P<0.5$ and $P>0.5$; P comparisons
transferred between $N=1$--15 and $N=16$--30. Each raw-response SVM used 800
training and 800 test pairs balanced by congruency, training within one
domain and testing in the other. Scores averaged the two transfer directions.

\subsection{Interventions on representational geometry}
\label{app:geometry_interventions}
To test whether geometry itself affects readout, we edited the responses of
frozen networks, leaving images and weights unchanged. We used one pretrained
checkpoint each of AlexNet, ResNet-50, ResNeXt-50, EfficientNet-B0,
DenseNet121, ViT-B/16, Swin-T, DeiT III-L/16, UniCL Swin-T, gMLP-S, Vim-S, and
PoolFormer-S24: relu5 responses for AlexNet and the final pooled or 
CLS features for the other models, with the subspace dimensions of Section~\ref{app:angle}.
The 18,000 stimuli were split into 10,039 training,
2,537 validation, and 5,424 test images. Subspaces were fitted as in
Section~\ref{app:angle}, but on the training split, and $Q$ denotes an
orthonormal basis of their joint span (dimension $d_N+d_P$; three for
AlexNet). Both edits were confined to this span.

\paragraph{Principal-angle manipulation.}
We adapted embedded deformation \citep{sumner2007embedded} with
as-rigid-as-possible regularization \citep{sorkine2007arap}. In coordinates
$\mathbf z=(Q^\top\mathbf x-\mathbf m)/\rho$, normalized by the training
center $\mathbf m$ and scale $\rho$, rotations $R_k$ and translations
$\mathbf t_k$ attached to control points $\mathbf g_k$ were blended with
normalized Gaussian weights $w_k$:
\begin{equation}
f_\delta(\mathbf z)=\mathbf z+\sum_k w_k(\mathbf z)
 \left[(R_k-I)(\mathbf z-\mathbf g_k)+\mathbf t_k\right]-\bar{\mathbf d},
\qquad
\mathbf x^{(\delta)}=\mathbf x+\rho\,Q\left[f_\delta(\mathbf z)-\mathbf z\right],
\label{eq:angle-edit}
\end{equation}
where $\bar{\mathbf d}$, the mean displacement of the training responses,
keeps the training mean fixed. Penalties on non-rigid local motion and on
changes in the size and aspect ratio of condition covariances limited
distortion. Maps were fitted on training and validation responses so that the
smallest training principal angle changed by $\delta$, from $-15^\circ$ to
$+15^\circ$ in $5^\circ$ steps (with any further principal angles held fixed),
and were applied unchanged to test responses. Angles were recomputed from the
edited full-dimensional responses, and no overlap correction was applied. The
$-15^\circ$ target was infeasible for DenseNet121, whose original angle was
below $15^\circ$, and was omitted.

\paragraph{Overlap manipulation.}
Within-cell variability was rescaled around the mean
$\boldsymbol\mu_{c,s}$ of each cell $c$, computed separately in each split
$s$:
\begin{equation}
\mathbf x^{(a)}=\mathbf x+(a-1)\,QQ^\top(\mathbf x-\boldsymbol\mu_{c,s}),
\qquad a\in\{0.25,0.5,1,2,4\}.
\label{eq:overlap-edit}
\end{equation}
This multiplies the within-cell covariance in the joint span by $a^2$ and
leaves cell means, and hence quantity subspaces and principal angles,
unchanged. Because the edit uses the cell labels of test images, it is an
intervention on representations, not a transformation applicable to unlabeled
images.

\paragraph{Evaluation.}
At every level of either manipulation, overlap was the mean Gaussian BC
(Eq.~\ref{eq:bc}) over all pairs of test cells, with Ledoit--Wolf
covariances \citep{ledoit2004well} estimated in the fixed frame $Q$. Affine
softmax decoders of N and P were refitted to the edited responses (AdamW,
weight decay $10^{-5}$, batch size 128, validation-loss early stopping);
accuracy was their test accuracy before ablation. Damage followed
Eqs.~\ref{eq:ablation-full} and~\ref{eq:damage}, holding the refitted decoder
fixed, with $\boldsymbol\mu$ the edited training mean and, for the angle
manipulation, subspaces re-estimated from the edited training responses; N
decoding removed the P subspace, and P decoding removed the N subspace.
Each value was divided by the same architecture's value for the original
representation, and the ratios were averaged across architectures with equal
weight (11 architectures at $-15^\circ$, 12 elsewhere); the plotted angle
change is $\delta/15^\circ$. With one checkpoint per architecture, these
summaries are descriptive. The two manipulations used separately fitted
decoders, so their original values differ, and comparisons are made within
each manipulation.

\subsection{Cross-domain ablation and damage}
\label{app:ablation}
Cross-domain ablation tests how strongly a fixed readout of one quantity
relies on the other quantity's subspace. The network, decoders, subspace bases
and train/test split were held fixed; only the relu5 responses passed to the
decoder were modified. Let $\mathbf x\in\mathbb R^{D}$ ($D=43{,}264$) be the
flattened relu5 response to one stimulus, $V_P\in\mathbb R^{D\times2}$ the
orthonormal basis of P2 and $\mathbf u_N$ the unit vector spanning N1, both
fitted on the full evaluation set, and $\boldsymbol\mu$ the mean response of
that network over its 14,400 decoder-training stimuli. Ablation removed from
each response the component of its deviation from $\boldsymbol\mu$ lying in
the opposite quantity's subspace:
\begin{equation}
\mathbf x_{\setminus P}=\mathbf x-V_PV_P^\top(\mathbf x-\boldsymbol\mu),\qquad
\mathbf x_{\setminus N}=\mathbf x-\mathbf u_N\mathbf u_N^\top(\mathbf x-\boldsymbol\mu).
\label{eq:ablation-full}
\end{equation}
All stimuli thus share the mean's coordinates within the removed subspace
(e.g., $V_P^\top\mathbf x_{\setminus P}=V_P^\top\boldsymbol\mu$), while their
coordinates in its orthogonal complement are unchanged: stimulus-dependent
variation along the removed directions is eliminated without shifting the
responses as a whole relative to the fixed decoder.

Each decoder was a linear classifier,
$\widehat y_q(\mathbf x)=\arg\max_c\,(W_q\mathbf x+\mathbf b_q)_c$, trained
once on intact training responses; neither $W_q$ nor $\mathbf b_q$ was
refitted after ablation. Damage N removed P2 and evaluated the fixed N
decoder; Damage P removed N1 and evaluated the fixed P decoder. For the
held-out test set $T$ (1,800 stimuli) and opposite quantity $\bar q$, damage
was
\begin{equation}
\Delta\mathrm{MAE}_q=\frac{1}{|T|}\sum_{i\in T}
 \left(|\widehat y_q(\mathbf x_{i,\setminus\bar q})-y_{qi}|
       -|\widehat y_q(\mathbf x_i)-y_{qi}|\right).
\label{eq:damage}
\end{equation}
Here $\widehat y_q$ is the predicted and $y_{qi}$ the true class index, with
both terms evaluated on the same stimuli. Positive damage denotes increased
error, measured in N steps or P-bin steps. Because the decoder is fixed,
damage quantifies the reliance of that readout on the removed subspace, not
whether the quantity could be recovered by a refitted decoder.

To ensure that ablation-induced damage was evaluated only when baseline decoding performance was meaningful, we established an explicit inclusion criterion using a constant-median predictor that always outputs the median target class without using the image. For each quantity, we included a network state in the principal-angle–ablation-damage correlation analysis only if its pre-ablation MAE was lower than that of the corresponding constant predictor. Under this criterion, the Label-shuffled and Noise-SSL conditions were excluded from the correlation analysis (Figure S3d,e). Their ablation results are nevertheless reported in Appendix Table 3 for completeness.

As a control, we also performed mean-preserving ablation of randomly selected subspaces matched in dimensionality to the removed quantity subspace. We evaluated the resulting decoding damage using the same fixed decoder and ablation procedure. This comparison allowed us to assess whether the observed damage was specific to the quantity-related subspace rather than a nonspecific effect of removing an equally dimensional subspace (Figure S9).

\subsection{Replication and statistical analysis}
Network initializations were the independent replication units. Geometry and
tasks from AlexNet were checkpoint-matched; folds and repeated unit draws were averaged
within networks. Tables show average and endpoint geometry maps show sample SD, while
paired comparison bars show SEM. Untrained--Pretrained comparisons used
two-sided paired $t$-tests across 20 matched initializations; the introductory
comparison baseline also used a one-sided test against 50\% chance.
Cross-state Pearson correlations and two-sided tests used equally weighted
state means, with sample size equal to the included state count. Trend lines
show ordinary least-squares fits with 95\% confidence intervals; state
inclusion followed each analysis.

\clearpage
\section{Complete task performance}
\label{app:performance}
Tables~\ref{tab:function-decodability}--\ref{tab:function-robustness} report functional
outcomes across architectures and AlexNet training states: decodability of numerosity and proportion
(Table~\ref{tab:function-decodability}, Figure~\ref{fig:supp-14}), comparison under conflicting quantity cues
(Table~\ref{tab:function-conflict}), and range generalization and cross-domain ablation
(Table~\ref{tab:function-robustness}). The pairwise comparison protocols and the direction of each
ablation are defined in Appendix~\ref{app:function} and Appendix~\ref{app:ablation}.
Table~\ref{tab:geometry-appendix} reports the representational geometry of the
same network states; estimation procedures are given in
Appendix~\ref{app:geometry}.
\begin{center}
\centering
\captionof{table}{Decodability of numerosity and proportion across architectures and AlexNet training states.}
\label{tab:function-decodability}
\begingroup\small
\setlength{\tabcolsep}{3pt}
\renewcommand{\arraystretch}{1.10}
\resizebox{\ifdim\width>\textwidth\textwidth\else\width\fi}{!}{%
\begin{tabular}{lccc!{\color{gray!55}\vrule width 0.5pt}cc}
\toprule
 & \multicolumn{3}{c}{Linear decoding (\%)} & \multicolumn{2}{c}{Pairwise comparison (\%)} \\
\cmidrule(lr){2-4}\cmidrule(lr){5-6}
 & N & P & NP & N & P \\
\midrule
\multicolumn{6}{l}{\textit{Untrained / Pretrained}} \\
AlexNet & $21.4$ / $\mathbf{29.2}$ & $69.2$ / $\mathbf{82.9}$ & $10.4$ / $\mathbf{17.5}$ & $93.2$ / $\mathbf{95.8}$ & $90.0$ / $\mathbf{90.3}$ \\
ResNet-50 & $16.6$ / $\mathbf{27.6}$ & $31.5$ / $\mathbf{78.6}$ & $5.4$ / $\mathbf{19.9}$ & $90.8$ / $\mathbf{96.2}$ & $81.3$ / $\mathbf{94.0}$ \\
ResNeXt-50 & $6.2$ / $\mathbf{27.7}$ & $33.3$ / $\mathbf{75.4}$ & $2.2$ / $\mathbf{21.7}$ & $78.8$ / $\mathbf{95.4}$ & $82.6$ / $\mathbf{96.4}$ \\
EfficientNet-B0 & $3.4$ / $\mathbf{29.8}$ & $15.4$ / $\mathbf{81.2}$ & $1.7$ / $\mathbf{23.7}$ & $58.9$ / $\mathbf{96.0}$ & $69.0$ / $\mathbf{98.0}$ \\
DenseNet121 & $11.8$ / $\mathbf{29.6}$ & $45.4$ / $\mathbf{67.9}$ & $3.9$ / $\mathbf{18.0}$ & $\mathbf{97.4}$ / $96.8$ & $\mathbf{93.5}$ / $93.2$ \\
ViT-B/16 & $5.9$ / $\mathbf{31.4}$ & $62.4$ / $\mathbf{80.7}$ & $4.4$ / $\mathbf{23.3}$ & $96.8$ / $\mathbf{98.4}$ & $92.1$ / $\mathbf{97.2}$ \\
Swin-T & $5.7$ / $\mathbf{30.1}$ & $38.2$ / $\mathbf{75.9}$ & $2.2$ / $\mathbf{23.1}$ & $86.5$ / $\mathbf{96.4}$ & $89.6$ / $\mathbf{95.5}$ \\
DeiT III-L/16 & $3.2$ / $\mathbf{27.7}$ & $26.4$ / $\mathbf{77.1}$ & $2.2$ / $\mathbf{23.8}$ & $73.8$ / $\mathbf{97.6}$ & $69.4$ / $\mathbf{96.4}$ \\
UniCL Swin-T & $6.2$ / $\mathbf{30.2}$ & $38.9$ / $\mathbf{54.6}$ & $2.2$ / $\mathbf{9.7}$ & $84.2$ / $\mathbf{96.9}$ & $91.2$ / $\mathbf{92.6}$ \\
gMLP-S & $8.2$ / $\mathbf{31.0}$ & $48.7$ / $\mathbf{61.6}$ & $3.8$ / $\mathbf{18.3}$ & $95.4$ / $\mathbf{96.4}$ & $92.2$ / $\mathbf{92.8}$ \\
Vim-S & $5.1$ / $\mathbf{28.6}$ & $21.4$ / $\mathbf{71.7}$ & $2.1$ / $\mathbf{19.2}$ & $64.6$ / $\mathbf{95.5}$ & $56.6$ / $\mathbf{93.9}$ \\
PoolFormer-S24 & $3.5$ / $\mathbf{34.9}$ & $35.2$ / $\mathbf{66.6}$ & $3.0$ / $\mathbf{19.1}$ & $67.1$ / $\mathbf{96.2}$ & $82.1$ / $\mathbf{93.2}$ \\
\midrule
\multicolumn{6}{l}{\textit{AlexNet, other training states}} \\
DVD & $25.4$ & $83.2$ & $15.5$ & $95.7$ & $90.2$ \\
Label-shuffled & $4.9$ & $23.9$ & $1.1$ & $68.9$ & $65.0$ \\
SimCLR & $21.4$ & $78.7$ & $12.3$ & $96.7$ & $90.9$ \\
RotNet & $25.1$ & $74.7$ & $14.6$ & $93.3$ & $89.7$ \\
Noise & $5.9$ & $27.5$ & $2.9$ & $65.8$ & $71.2$ \\
Grating & $18.7$ & $57.1$ & $9.7$ & $89.0$ & $86.6$ \\
Statistical & $21.1$ & $77.8$ & $12.4$ & $96.4$ & $92.0$ \\
Shaders & $20.1$ & $81.7$ & $12.5$ & $96.7$ & $90.5$ \\
\bottomrule
\end{tabular}}
\par\medskip
\begin{minipage}{\textwidth}\footnotesize
Upper rows give Untrained / Pretrained values for each architecture, with the better value in bold (higher accuracy). Lower rows give AlexNet under the other training conditions. Values are means across networks: 20 for AlexNet Untrained, Pretrained, DVD, and Label-shuffled, 5 for the other AlexNet states and for ResNet-50, and one for the other architectures. Linear N, P, and NP denote 30-, 7-, and 195-class decoding accuracy from the full representation. Pairwise comparison is the standard N/P comparison (400 N/P-congruent and 400 N/P-incongruent test pairs) with linear SVMs on raw concatenated pair features ($C=1$, hinge loss, no scaling or PCA). PoolFormer-S24 uses the original implementation with normalization before pooling. Protocols are detailed in Appendix~\ref{app:function}.
\end{minipage}
\endgroup
\end{center}

\clearpage
\begin{center}
\centering
\captionof{table}{Comparison under conflicting quantity cues across architectures and AlexNet training states.}
\label{tab:function-conflict}
\begingroup\small
\setlength{\tabcolsep}{3pt}
\renewcommand{\arraystretch}{1.10}
\resizebox{\ifdim\width>\textwidth\textwidth\else\width\fi}{!}{%
\begin{tabular}{lcc!{\color{gray!55}\vrule width 0.5pt}cc!{\color{gray!55}\vrule width 0.5pt}cc}
\toprule
 & \multicolumn{2}{c}{\shortstack{P comparison (\%)\\$N_W$ conflict}} & \multicolumn{2}{c}{\shortstack{N comparison (\%)\\P conflict}} & \multicolumn{2}{c}{\shortstack{P comparison (\%)\\N conflict}} \\
\cmidrule(lr){2-3}\cmidrule(lr){4-5}\cmidrule(lr){6-7}
 & Congruent & Incongruent & Congruent & Incongruent & Congruent & Incongruent \\
\midrule
\multicolumn{7}{l}{\textit{Untrained / Pretrained}} \\
AlexNet & $76.9$ / $\mathbf{78.9}$ & $49.9$ / $\mathbf{60.6}$ & $57.4$ / $\mathbf{71.3}$ & $58.4$ / $\mathbf{68.1}$ & $60.0$ / $\mathbf{67.5}$ & $51.6$ / $\mathbf{61.2}$ \\
ResNet-50 & $71.2$ / $\mathbf{89.4}$ & $49.9$ / $\mathbf{82.4}$ & $56.7$ / $\mathbf{76.2}$ & $56.2$ / $\mathbf{71.7}$ & $56.2$ / $\mathbf{73.0}$ & $51.2$ / $\mathbf{71.2}$ \\
ResNeXt-50 & $75.9$ / $\mathbf{93.3}$ & $40.6$ / $\mathbf{86.2}$ & $55.1$ / $\mathbf{67.5}$ & $49.5$ / $\mathbf{66.9}$ & $62.0$ / $\mathbf{67.5}$ & $42.8$ / $\mathbf{76.3}$ \\
EfficientNet-B0 & $78.4$ / $\mathbf{96.6}$ & $37.5$ / $\mathbf{92.5}$ & $72.6$ / $\mathbf{76.2}$ & $31.7$ / $\mathbf{76.3}$ & $65.5$ / $\mathbf{79.3}$ & $43.0$ / $\mathbf{82.3}$ \\
DenseNet121 & $\mathbf{90.8}$ / $87.7$ & $55.0$ / $\mathbf{78.8}$ & $80.3$ / $\mathbf{82.4}$ & $76.3$ / $\mathbf{79.2}$ & $\mathbf{72.0}$ / $69.0$ & $63.3$ / $\mathbf{71.0}$ \\
ViT-B/16 & $87.2$ / $\mathbf{98.3}$ & $61.9$ / $\mathbf{95.0}$ & $67.8$ / $\mathbf{87.6}$ & $74.5$ / $\mathbf{86.3}$ & $68.8$ / $\mathbf{93.8}$ & $62.5$ / $\mathbf{96.0}$ \\
Swin-T & $75.2$ / $\mathbf{94.8}$ & $60.6$ / $\mathbf{86.2}$ & $53.5$ / $\mathbf{77.9}$ & $53.3$ / $\mathbf{74.4}$ & $50.5$ / $\mathbf{79.3}$ & $63.8$ / $\mathbf{80.0}$ \\
DeiT III-L/16 & $50.0$ / $\mathbf{94.7}$ & $50.0$ / $\mathbf{90.0}$ & $50.0$ / $\mathbf{83.0}$ & $50.0$ / $\mathbf{80.1}$ & $50.0$ / $\mathbf{86.3}$ & $50.0$ / $\mathbf{84.0}$ \\
UniCL Swin-T & $74.5$ / $\mathbf{91.4}$ & $58.8$ / $\mathbf{61.3}$ & $53.6$ / $\mathbf{80.0}$ & $51.6$ / $\mathbf{72.6}$ & $50.8$ / $\mathbf{65.5}$ & $\mathbf{64.3}$ / $62.8$ \\
gMLP-S & $\mathbf{88.3}$ / $88.0$ & $44.4$ / $\mathbf{79.4}$ & $66.6$ / $\mathbf{77.1}$ & $67.8$ / $\mathbf{77.8}$ & $\mathbf{70.3}$ / $57.8$ & $60.3$ / $\mathbf{64.8}$ \\
Vim-S & $57.3$ / $\mathbf{83.8}$ & $48.8$ / $\mathbf{76.9}$ & $49.6$ / $\mathbf{70.1}$ & $51.0$ / $\mathbf{66.5}$ & $50.0$ / $\mathbf{63.0}$ & $48.3$ / $\mathbf{72.8}$ \\
PoolFormer-S24 & $72.2$ / $\mathbf{89.4}$ & $60.6$ / $\mathbf{74.4}$ & $49.0$ / $\mathbf{82.6}$ & $51.8$ / $\mathbf{82.0}$ & $50.8$ / $\mathbf{72.8}$ & $59.0$ / $\mathbf{77.0}$ \\
\midrule
\multicolumn{7}{l}{\textit{AlexNet, other training states}} \\
DVD & $78.9$ & $59.8$ & $66.8$ & $63.1$ & $68.2$ & $62.9$ \\
Label-shuffled & $54.6$ & $59.3$ & $44.8$ & $56.8$ & $44.7$ & $56.2$ \\
SimCLR & $82.8$ & $61.6$ & $68.0$ & $68.4$ & $69.7$ & $61.7$ \\
RotNet & $74.8$ & $52.6$ & $61.4$ & $61.8$ & $63.5$ & $57.2$ \\
Noise & $52.3$ & $72.5$ & $45.5$ & $58.2$ & $44.0$ & $60.7$ \\
Grating & $70.5$ & $65.9$ & $54.3$ & $59.4$ & $59.0$ & $57.3$ \\
Statistical & $84.4$ & $65.9$ & $65.7$ & $64.4$ & $71.9$ & $63.3$ \\
Shaders & $82.9$ & $59.6$ & $67.7$ & $72.4$ & $70.5$ & $62.9$ \\
\bottomrule
\end{tabular}}
\par\medskip
\begin{minipage}{\textwidth}\footnotesize
Upper rows give Untrained / Pretrained values for each architecture, with the better value in bold (higher accuracy). Lower rows give AlexNet under the other training conditions. Values are means across networks: 20 for AlexNet Untrained, Pretrained, DVD, and Label-shuffled, 5 for the other AlexNet states and for ResNet-50, and one for the other architectures. Accuracy on congruent and incongruent test pairs of three pairwise comparison tasks (linear SVMs as in Table~\ref{tab:function-decodability}). P comparison under $N_W$ conflict: pairs one P bin apart, with congruency defined by white-dot count $N_W$ (640 congruent and 160 incongruent test pairs). N comparison under P conflict: pairs with $|N_a-N_b|=1$. P comparison under N conflict: pairs with $0.045\leq|P_a-P_b|\leq0.055$. In the last two tasks, the restriction applies to both the 800 training and the 800 test pairs, and congruency is defined by N and P. Protocols are detailed in Appendix~\ref{app:function}.
\end{minipage}
\endgroup
\end{center}

\clearpage
\begin{center}
\centering
\captionof{table}{Cross-quantity robustness across architectures and AlexNet training states: range generalization and cross-domain ablation. Gray values mark states whose decoder is not better than a constant predictor.}
\label{tab:function-robustness}
\begingroup\small
\setlength{\tabcolsep}{3pt}
\renewcommand{\arraystretch}{1.10}
\resizebox{\ifdim\width>\textwidth\textwidth\else\width\fi}{!}{%
\begin{tabular}{lcc!{\color{gray!55}\vrule width 0.5pt}cc!{\color{gray!55}\vrule width 0.5pt}cc}
\toprule
 & \multicolumn{2}{c}{Generalization (\%)} & \multicolumn{2}{c}{Baseline MAE} & \multicolumn{2}{c}{Damage ($\Delta$MAE)} \\
\cmidrule(lr){2-3}\cmidrule(lr){4-5}\cmidrule(lr){6-7}
 & N & P & N & P & N & P \\
\midrule
\multicolumn{7}{l}{\textit{Untrained / Pretrained}} \\
AlexNet & $85.8$ / $\mathbf{92.0}$ & $86.3$ / $\mathbf{87.9}$ & $1.85$ / $\mathbf{1.27}$ & $0.32$ / $\mathbf{0.17}$ & $3.22$ / $\mathbf{1.97}$ & $\mathbf{0.26}$ / $0.34$ \\
ResNet-50 & $86.1$ / $\mathbf{91.1}$ & $72.2$ / $\mathbf{89.1}$ & $2.70$ / $\mathbf{1.52}$ & $1.34$ / $\mathbf{0.22}$ & $6.38$ / $\mathbf{1.17}$ & $0.33$ / $\mathbf{0.20}$ \\
ResNeXt-50 & $72.3$ / $\mathbf{88.3}$ & $79.1$ / $\mathbf{92.4}$ & \textcolor{black!45}{$10.21$} / $\mathbf{1.66}$ & $1.28$ / $\mathbf{0.25}$ & \textcolor{black!45}{$3.76$} / $0.93$ & $\mathbf{-0.12}$ / $0.06$ \\
EfficientNet-B0 & $61.8$ / $\mathbf{89.3}$ & $73.3$ / $\mathbf{90.8}$ & \textcolor{black!45}{$9.36$} / $\mathbf{1.39}$ & \textcolor{black!45}{$1.81$} / $\mathbf{0.19}$ & \textcolor{black!45}{$0.00$} / $0.85$ & \textcolor{black!45}{$0.00$} / $0.03$ \\
DenseNet121 & $94.5$ / $\mathbf{95.1}$ & $\mathbf{89.6}$ / $86.1$ & $4.34$ / $\mathbf{1.42}$ & $0.97$ / $\mathbf{0.38}$ & $7.70$ / $\mathbf{2.04}$ & $\mathbf{0.02}$ / $0.09$ \\
ViT-B/16 & $89.6$ / $\mathbf{90.4}$ & $\mathbf{90.9}$ / $89.9$ & $5.18$ / $\mathbf{1.40}$ & $0.43$ / $\mathbf{0.20}$ & $\mathbf{-0.92}$ / $1.28$ & $0.20$ / $\mathbf{0.07}$ \\
Swin-T & $75.8$ / $\mathbf{91.0}$ & $\mathbf{86.4}$ / $84.1$ & \textcolor{black!45}{$8.18$} / $\mathbf{1.44}$ & $0.83$ / $\mathbf{0.26}$ & \textcolor{black!45}{$0.49$} / $0.21$ & $\mathbf{0.04}$ / $0.13$ \\
DeiT III-L/16 & $66.6$ / $\mathbf{89.4}$ & $68.1$ / $\mathbf{88.1}$ & \textcolor{black!45}{$13.50$} / $\mathbf{1.53}$ & $1.47$ / $\mathbf{0.24}$ & \textcolor{black!45}{$0.08$} / $0.96$ & $0.35$ / $\mathbf{0.06}$ \\
UniCL Swin-T & $74.0$ / $\mathbf{93.2}$ & $\mathbf{87.0}$ / $84.4$ & \textcolor{black!45}{$8.52$} / $\mathbf{1.63}$ & $0.83$ / $\mathbf{0.66}$ & \textcolor{black!45}{$2.93$} / $2.02$ & $0.10$ / $\mathbf{0.06}$ \\
gMLP-S & $\mathbf{93.2}$ / $90.9$ & $\mathbf{91.7}$ / $81.0$ & $5.27$ / $\mathbf{1.49}$ & $0.81$ / $\mathbf{0.52}$ & $2.21$ / $\mathbf{0.72}$ & $\mathbf{-0.12}$ / $0.02$ \\
Vim-S & $57.9$ / $\mathbf{89.3}$ & $56.2$ / $\mathbf{81.6}$ & \textcolor{black!45}{$8.42$} / $\mathbf{1.46}$ & \textcolor{black!45}{$1.94$} / $\mathbf{0.34}$ & \textcolor{black!45}{$-0.17$} / $1.32$ & \textcolor{black!45}{$-0.22$} / $0.10$ \\
PoolFormer-S24 & $56.9$ / $\mathbf{93.9}$ & $\mathbf{83.9}$ / $80.0$ & \textcolor{black!45}{$13.12$} / $\mathbf{1.18}$ & $1.01$ / $\mathbf{0.41}$ & \textcolor{black!45}{$0.59$} / $0.28$ & $\mathbf{-0.08}$ / $0.10$ \\
\midrule
\multicolumn{7}{l}{\textit{AlexNet, other training states}} \\
DVD & $89.0$ & $86.9$ & $1.52$ & $0.17$ & $0.77$ & $0.37$ \\
Label-shuffled & $62.5$ & $67.1$ & \textcolor{black!45}{$9.20$} & \textcolor{black!45}{$1.79$} & \textcolor{black!45}{$2.30$} & \textcolor{black!45}{$0.48$} \\
SimCLR & $88.6$ & $89.1$ & $1.81$ & $0.21$ & $1.13$ & $0.28$ \\
RotNet & $90.1$ & $88.8$ & $1.54$ & $0.27$ & $2.80$ & $0.36$ \\
Noise & $61.2$ & $73.4$ & \textcolor{black!45}{$8.45$} & $1.43$ & \textcolor{black!45}{$3.29$} & $0.55$ \\
Grating & $82.9$ & $80.1$ & $2.40$ & $0.49$ & $2.38$ & $0.15$ \\
Statistical & $85.8$ & $90.7$ & $1.84$ & $0.22$ & $1.31$ & $0.21$ \\
Shaders & $87.5$ & $89.3$ & $1.95$ & $0.18$ & $0.65$ & $0.36$ \\
\bottomrule
\end{tabular}}
\par\medskip
\begin{minipage}{\textwidth}\footnotesize
Upper rows give Untrained / Pretrained values for each architecture, with the better value in bold (higher accuracy; lower MAE or damage). Lower rows give AlexNet under the other training conditions. Values are means across networks: 20 for AlexNet Untrained, Pretrained, DVD, and Label-shuffled, 5 for the other AlexNet states and for ResNet-50, and one for the other architectures. Generalization averages both range-transfer directions. Baseline MAE is the error of the linear N or P decoder before ablation, in N steps or P-bin steps. Damage is $\mathrm{MAE}_{\mathrm{ablated}}-\mathrm{MAE}_{\mathrm{baseline}}$ after mean-preserving ablation, $x^{\prime}=x-UU^\top(x-\mu_{\mathrm{train}})$, where $\mu_{\mathrm{train}}$ is the decoder-training activation mean and the affine decoders remain fixed. For N, the P subspace was removed (rank 2); for P, the N subspace was removed (rank 1 for AlexNet, rank 2 for the other architectures). Gray values: the baseline MAE is not lower than that of a constant predictor that always answers the median class (7.48 for N, 1.65 for P; Figure~\ref{fig:supp-4}), so the decoder has no information to lose and damage is not interpretable; damage is compared (bold) only when both states are valid. Protocols are detailed in Appendix~\ref{app:ablation}.
\end{minipage}
\endgroup
\end{center}

\clearpage
\begin{center}
\centering
\captionof{table}{Representational geometry across architectures and AlexNet training states: subspace orientation, overlap, distance, and volume. AlexNet uses one N and two P components ($d_N=1$, $d_P=2$) at relu5; the other architectures use two of each ($d_N=d_P=2$) at the pooled or CLS feature.}
\label{tab:geometry-appendix}
\begingroup\small
\setlength{\tabcolsep}{2pt}
\renewcommand{\arraystretch}{1.10}
\resizebox{\ifdim\width>\textwidth\textwidth\else\width\fi}{!}{%
\begin{tabular}{lc!{\color{gray!55}\vrule width 0.5pt}ccc!{\color{gray!55}\vrule width 0.5pt}c!{\color{gray!55}\vrule width 0.5pt}c}
\toprule
 & PA ($^\circ$) & NP overlap & N overlap & P overlap & \shortstack{Normalized\\distance} & \shortstack{Normalized\\volume (95\%)} \\
\midrule
\multicolumn{7}{l}{\textit{Untrained / Pretrained}} \\
AlexNet & $5.8$ / $\mathbf{31.6}$ & $0.282$ / $\mathbf{0.083}$ & $0.547$ / $\mathbf{0.242}$ & $0.491$ / $\mathbf{0.216}$ & $92.8$ / $138.8$ & $5.28\times10^{5}$ / $4.69\times10^{5}$ \\
ResNet-50 & $0.4$ / $\mathbf{24.9}$ & $0.694$ / $\mathbf{0.109}$ & $0.706$ / $\mathbf{0.321}$ & $0.943$ / $\mathbf{0.260}$ & $30.4$ / $38.4$ & $6.79\times10^{5}$ / $4.50\times10^{5}$ \\
ResNeXt-50 & $1.3$ / $\mathbf{25.8}$ & $0.524$ / $\mathbf{0.137}$ & $0.784$ / $\mathbf{0.343}$ & $0.718$ / $\mathbf{0.284}$ & $32.7$ / $31.0$ & $7.68\times10^{6}$ / $2.47\times10^{5}$ \\
EfficientNet-B0 & $16.9$ / $\mathbf{29.9}$ & $0.532$ / $\mathbf{0.094}$ & $0.906$ / $\mathbf{0.306}$ & $0.755$ / $\mathbf{0.249}$ & $22.8$ / $25.5$ & $2.57\times10^{6}$ / $7.50\times10^{4}$ \\
DenseNet121 & $2.4$ / $\mathbf{11.7}$ & $0.534$ / $\mathbf{0.181}$ & $0.734$ / $\mathbf{0.306}$ & $0.720$ / $\mathbf{0.424}$ & $19.8$ / $21.8$ & $2.96\times10^{5}$ / $9.13\times10^{4}$ \\
ViT-B/16 & $17.8$ / $\mathbf{20.6}$ & $0.187$ / $\mathbf{0.073}$ & $0.679$ / $\mathbf{0.226}$ & $0.367$ / $\mathbf{0.312}$ & $28.2$ / $18.3$ & $2.07\times10^{5}$ / $3.42\times10^{3}$ \\
Swin-T & $22.3$ / $\mathbf{38.4}$ & $0.248$ / $\mathbf{0.087}$ & $0.715$ / $\mathbf{0.245}$ & $0.425$ / $\mathbf{0.267}$ & $19.4$ / $21.4$ & $2.43\times10^{5}$ / $2.59\times10^{4}$ \\
DeiT III-L/16 & $0.5$ / $\mathbf{21.0}$ & $0.261$ / $\mathbf{0.135}$ & $0.726$ / $\mathbf{0.362}$ & $0.439$ / $\mathbf{0.349}$ & $31.7$ / $18.0$ & $3.26\times10^{5}$ / $2.04\times10^{4}$ \\
UniCL Swin-T & $18.1$ / $\mathbf{18.8}$ & $0.245$ / $\mathbf{0.146}$ & $0.719$ / $\mathbf{0.251}$ & $\mathbf{0.417}$ / $0.449$ & $21.5$ / $17.3$ & $3.14\times10^{5}$ / $5.71\times10^{4}$ \\
gMLP-S & $0.7$ / $\mathbf{33.3}$ & $0.207$ / $\mathbf{0.154}$ & $0.559$ / $\mathbf{0.260}$ & $\mathbf{0.417}$ / $0.531$ & $15.5$ / $10.1$ & $9.79\times10^{3}$ / $3.01\times10^{3}$ \\
Vim-S & $9.8$ / $\mathbf{26.2}$ & $0.631$ / $\mathbf{0.123}$ & $0.864$ / $\mathbf{0.325}$ & $0.885$ / $\mathbf{0.348}$ & $6.5$ / $12.3$ & $1.91\times10^{5}$ / $8.14\times10^{3}$ \\
PoolFormer-S24 & $1.7$ / $\mathbf{39.8}$ & $0.352$ / $\mathbf{0.111}$ & $0.799$ / $\mathbf{0.208}$ & $0.575$ / $\mathbf{0.375}$ & $15.9$ / $16.1$ & $1.74\times10^{5}$ / $1.52\times10^{4}$ \\
\midrule
\multicolumn{7}{l}{\textit{AlexNet, other training states}} \\
DVD & $51.9$ & $0.084$ & $0.289$ & $0.191$ & $141.6$ & $4.21\times10^{5}$ \\
Label-shuffled & $1.8$ & $0.565$ & $0.889$ & $0.809$ & $118.3$ & $2.25\times10^{7}$ \\
SimCLR & $46.8$ & $0.065$ & $0.253$ & $0.168$ & $118.7$ & $2.06\times10^{5}$ \\
RotNet & $32.6$ & $0.102$ & $0.263$ & $0.281$ & $102.0$ & $1.91\times10^{5}$ \\
Noise & $1.6$ & $0.561$ & $0.915$ & $0.740$ & $134.3$ & $2.44\times10^{7}$ \\
Grating & $35.5$ & $0.211$ & $0.431$ & $0.444$ & $123.1$ & $1.12\times10^{6}$ \\
Statistical & $39.9$ & $0.062$ & $0.292$ & $0.148$ & $101.4$ & $1.07\times10^{5}$ \\
Shaders & $58.3$ & $0.052$ & $0.216$ & $0.145$ & $128.7$ & $1.10\times10^{5}$ \\
\bottomrule
\end{tabular}}
\par\medskip
\begin{minipage}{\textwidth}\footnotesize
Upper rows give Untrained / Pretrained values for each architecture, with the larger principal angle and the smaller overlap in bold. Lower rows give AlexNet under the other training conditions. Values are means across networks for AlexNet, means across five networks for ResNet-50, and single-network values for the other ten architectures. PA is the smallest principal angle. Overlaps are Bhattacharyya coefficients (BC; Eq.~\ref{eq:bc}). Overlaps, centroid distances, and volumes are averaged across three held-out folds. Centroid distance and volume are normalized within each network as $D/a$ and $V_d/a^d$ before averaging, where $a$ is the centered per-feature RMS of the readout features, $V_d$ is the 95\% Gaussian ellipsoid volume, and $d=d_N+d_P$ ($d=3$ for AlexNet, $d=4$ for the other architectures). Distance and volume should be compared within an architecture, not across architectures. PA and BC are invariant to this normalization. See Appendix~\ref{app:geometry} for geometry estimation procedures.
\end{minipage}
\endgroup
\end{center}

\clearpage
\section{Supplementary figures}
\label{app:figures}
These figures expand the stimulus design, subspace dimensionality, the
ablation procedure, baseline decodability, color conditions, comparison
under W/P conflict, the relation between N/P conflict and W/P incongruency,
visual controls, representational interventions, label-correspondence
controls, synthetic-image distributions, learning trajectories,
geometry--function comparisons, and additional architectures used in this study. Analysis definitions are provided in
Appendix~\ref{app:methods}.

\begin{figure}[!htbp]
\centering
\includegraphics[width=0.81\linewidth]{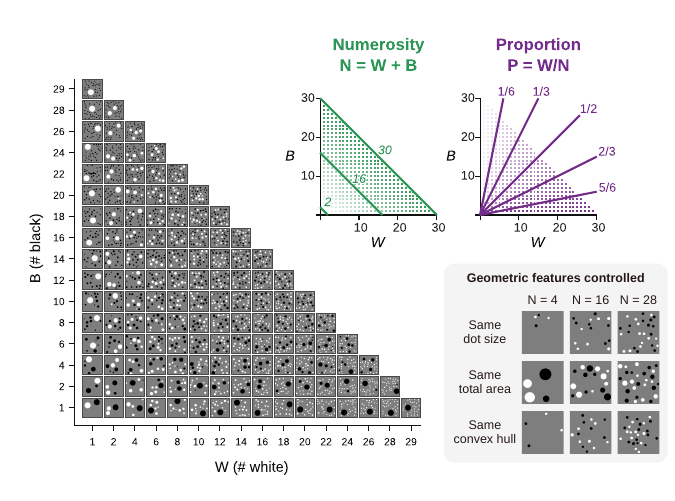}
\caption{Systematic variation of numerosity and proportion with geometric
controls. White-dot count $W$ and black-dot count $B$ jointly vary total
numerosity $N=W+B$ and proportion $P=W/N$ across the stimulus grid.
Constant-N and constant-P directions reveal the systematic organization of
the two quantities. Three stimulus regimes control dot size, total dot area,
and spatial extent, enabling quantity representations to be examined across
these geometric properties.}
\label{fig:supp-1}
\end{figure}

\clearpage
\begin{figure}[!htbp]
\centering
\includegraphics[width=1.00\linewidth]{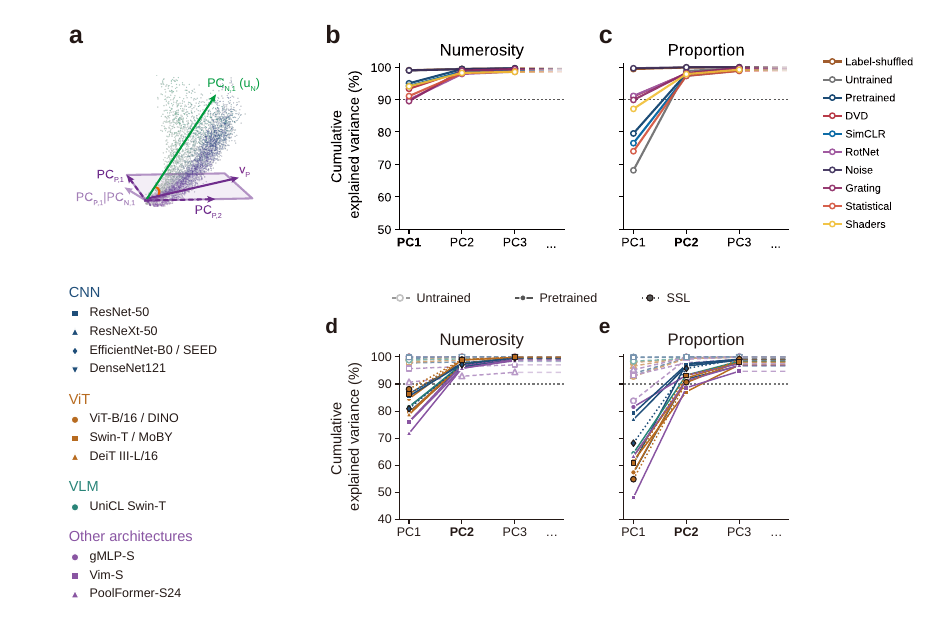}
\caption{Rationale for the subspace dimensions. (a) Schematic
of the one-dimensional numerosity subspace and two-dimensional proportion
subspace. (b,c) Cumulative explained variance of quantity-conditioned
responses for N and P in the AlexNet states. One N component and two P
components capture more than 90\% of the variance in most states (dotted
threshold). (d,e) The same analysis for the additional architectures in
their Untrained, Pretrained, and SSL states, with two components 
retained for both N and P (N2/P2); bold 
tick labels mark the number of components retained. Retaining these 
dimensions across states within each architecture provides a common 
criterion for comparing the two quantities and their geometry across 
network states.}
\label{fig:supp-2}
\end{figure}

\clearpage
\begin{figure}[!htbp]
\centering
\includegraphics[width=0.96\linewidth]{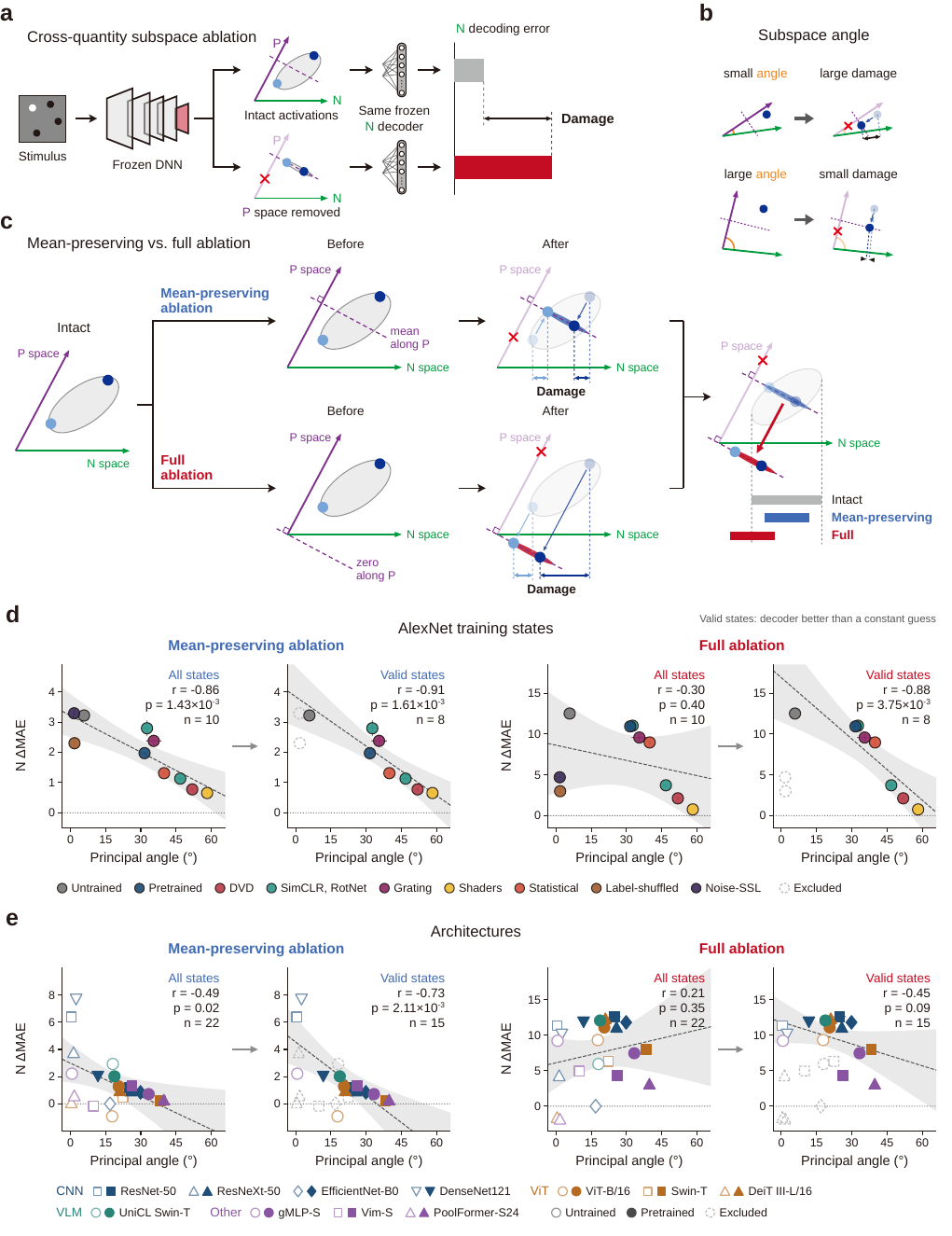}
\caption{Cross-quantity subspace ablation. (a) Activations of a frozen
network are read by the same frozen N decoder before and after the P
subspace is removed; damage is the increase in N decoding error.
(b) A smaller principal angle between the N and P subspaces gives a larger
displacement along N, and thus larger damage. (c) Mean-preserving ablation
replaces each activation's component along P with its mean (dashed line),
removing variation along P without shifting the activations; full ablation
sets the component to zero, shifting all activations outside the intact
range along N. (d, e) N damage ($\Delta$MAE) versus principal angle under
mean-preserving (left) and full (right) ablation, for AlexNet training
states (d) and additional architectures (e), using all states or only valid
states (Figure~\ref{fig:supp-4}). The same procedure is applied with N and
P exchanged.}
\label{fig:supp-3}
\end{figure}

\clearpage
\begin{figure}[!htbp]
\centering
\includegraphics[width=0.92\linewidth]{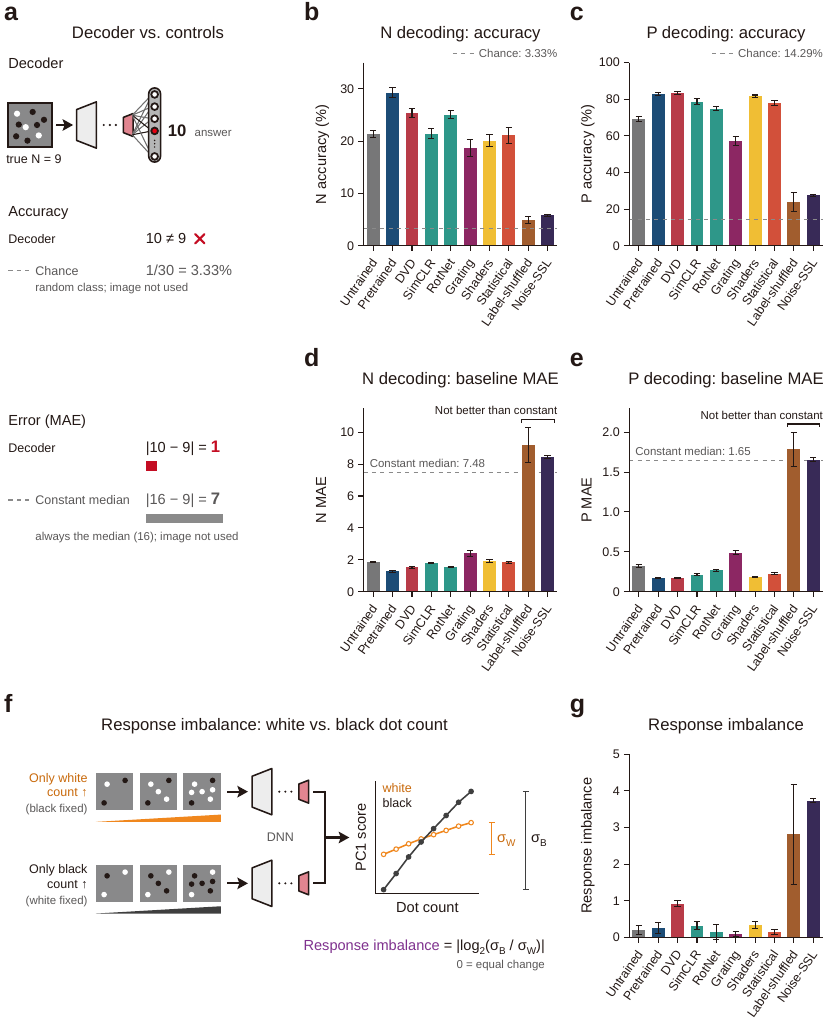}
\caption{Baseline decodability and response imbalance across network states.
(a) Reference levels: accuracy is compared with chance (a random class), and
error with a constant predictor that always answers the median class;
neither uses the image. (b,c) N and P decoding accuracy relative to chance
($1/30$ and $1/7$). (d,e) Baseline MAE relative to the constant-median
reference; Label-shuffled and Noise-SSL networks are not better than the
constant predictor. (f) Response imbalance: the effects of white-dot and
black-dot count on relu5 responses are each projected onto their first
principal component, and imbalance is $|\log_2(\sigma_B/\sigma_W)|$, where
$\sigma_W$ and $\sigma_B$ are the standard deviations of the PC1 scores
across count levels (0, equal change; curves are illustrative).
(g) Response imbalance in each state. Bars show mean $\pm$ SD across
networks (20 for Untrained, Pretrained, DVD, and Label-shuffled; 5 for the
other states; 3 for Noise-SSL, 4 in g).}
\label{fig:supp-4}
\end{figure}

\clearpage
\begin{figure}[!htbp]
\centering
\includegraphics[width=1.00\linewidth]{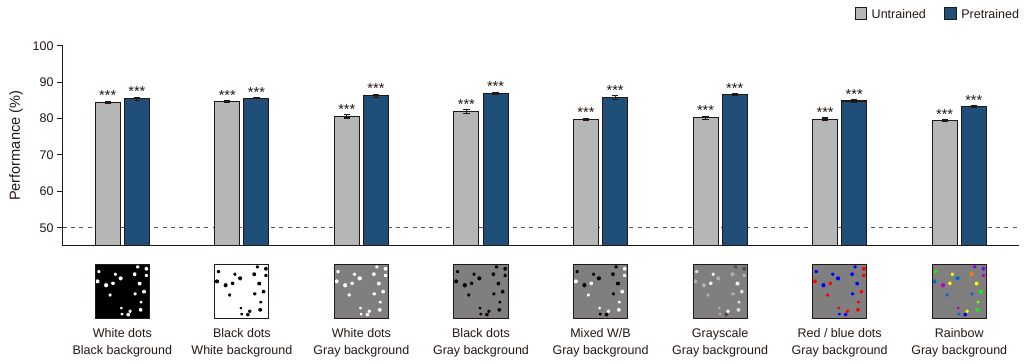}
\caption{Numerosity comparison across color conditions. Performance of
Untrained and Pretrained networks in the numerosity comparison task of
Figure~\ref{fig:main-2}b for eight combinations of dot and background colors
(examples below the bars). All conditions share dot layouts, dot sizes, image
splits, and comparison pairs. Units were reselected and the SVM was retrained
for each condition. Bars show mean $\pm$ SEM across five networks; the dashed
line indicates chance (50\%). $^{***}p<0.001$, one-sample $t$-test against
chance, Holm-corrected across all bars.}
\label{fig:supp-5}
\end{figure}

\clearpage
\begin{figure}[!htbp]
\centering
\includegraphics[width=1.00\linewidth]{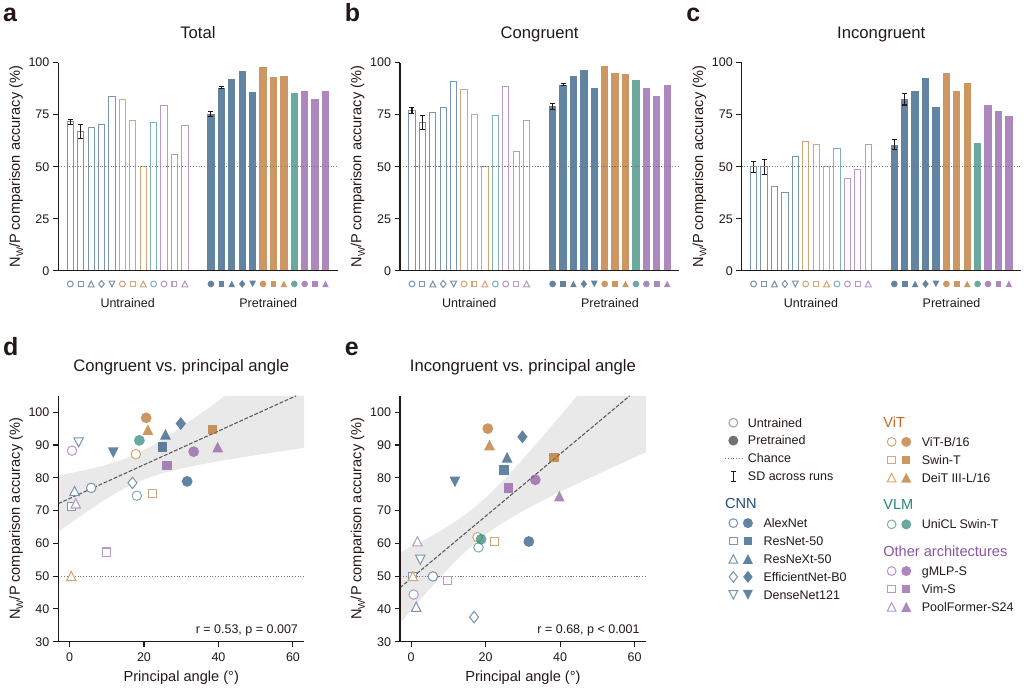}
\caption{Hard $N_W$/P comparison across architectures. A linear SVM compares
P between two images that are one P bin apart. (a--c) Accuracy on all,
congruent, and incongruent test pairs for the Untrained and Pretrained states
of 12 architectures; marker shapes and colors identify the models. Error bars
show SD across runs where several runs were available (AlexNet and
ResNet-50); dotted lines indicate chance (50\%).
(d,e) Congruent and incongruent accuracy against the smallest principal angle
between the N and P subspaces. Dashed lines show least-squares fits with 95\%
confidence bands, and $r$ is the Pearson correlation across the 24 states.}
\label{fig:supp-6}
\end{figure}

\clearpage
\begin{figure}[!htbp]
\centering
\includegraphics[width=1.00\linewidth]{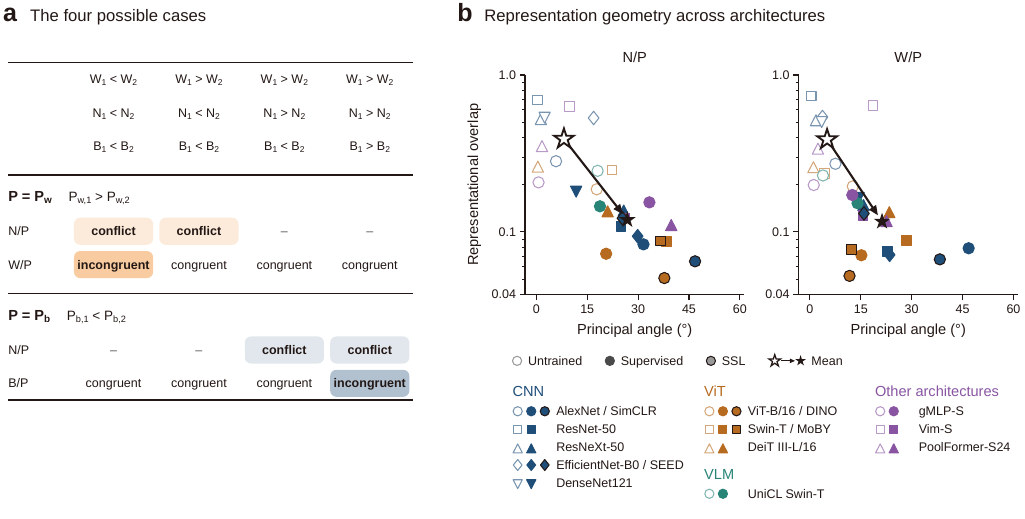}
\caption{N/P conflict contains W/P and B/P incongruency. For a pair of
images, a count is congruent with a proportion when it is larger in the image
with the larger proportion, and incongruent otherwise; for the total number
$N$ we refer to this as conflict. Image 1 denotes the image with the larger
proportion of white dots ($P_{w,1}>P_{w,2}$, equivalently
$P_{b,1}<P_{b,2}$). (a) The four possible orderings of $W$, $N$, and $B$.
With $P=P_w$, N conflicts with P when $N_1<N_2$, and the W/P-incongruent
pairs ($W_1<W_2$) are a subset of these pairs. With $P=P_b$, the conflict
side reverses ($N_1>N_2$) and contains the B/P-incongruent pairs
($B_1>B_2$). (b) Principal angle and representational overlap across
architectures for the N and P subspaces (left) and for the W and P subspaces
(right). Stars show the means of the 12 untrained and the 12 supervised
networks.}
\label{fig:supp-7}
\end{figure}

\clearpage
\begin{figure}[!htbp]
\centering
\includegraphics[width=1.00\linewidth]{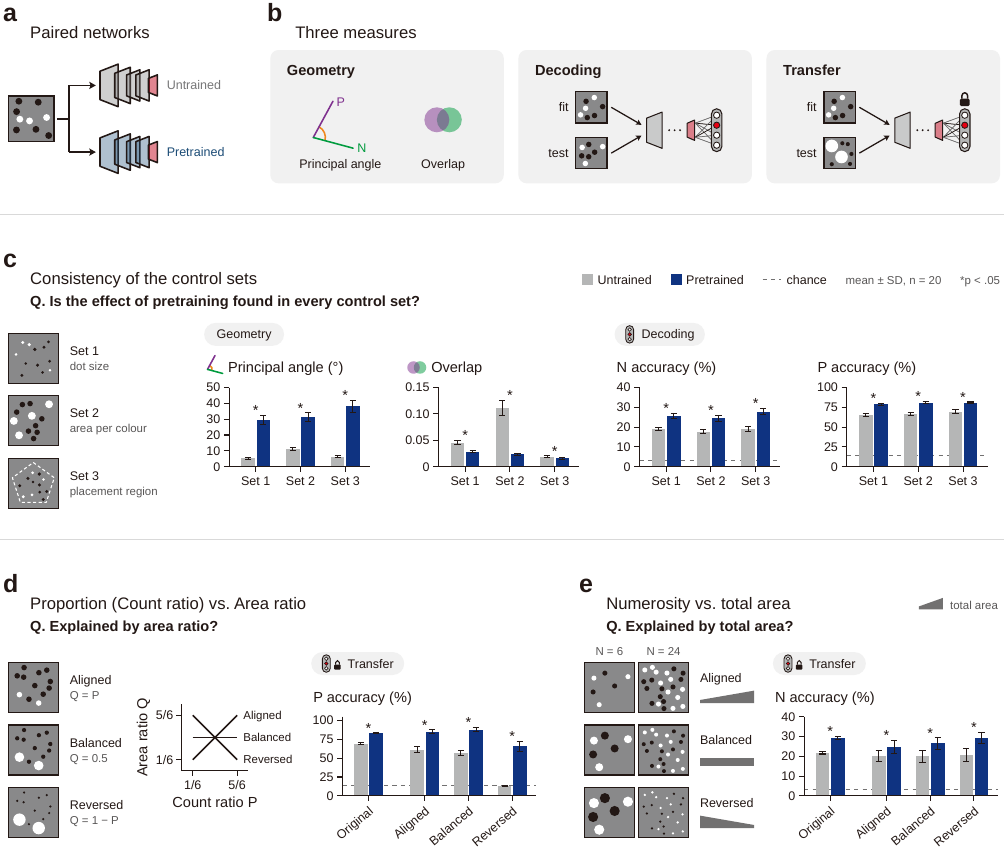}
\caption{Visual controls for the effect of pretraining. (a) Untrained
networks are compared with their Pretrained counterparts (20 pairs).
(b) Three measures: geometry (principal angle and overlap), decoding (a
linear readout fitted and tested within a stimulus set), and transfer (a
readout fitted on the original stimuli, frozen, and tested on new stimuli).
(c) In each stimulus regime of Figure~\ref{fig:supp-1}, pretraining increases
the principal angle, reduces overlap, and improves N and P decoding.
(d) The frozen P readout of Pretrained networks remains more accurate when
the white-area ratio $Q$ is aligned with, balanced against, or reversed
relative to the count ratio $P$. (e) The frozen N readout remains more
accurate when total dot area increases with, is balanced across, or
decreases with $N$ (example images exaggerate the difference). Bars show
mean $\pm$ SD across networks; dashed lines indicate chance. $^{*}p<0.05$,
paired comparison of Untrained and Pretrained networks, Holm-corrected.}
\label{fig:supp-8}
\end{figure}

\clearpage
\begin{figure}[!htbp]
\centering
\includegraphics[width=1.00\linewidth]{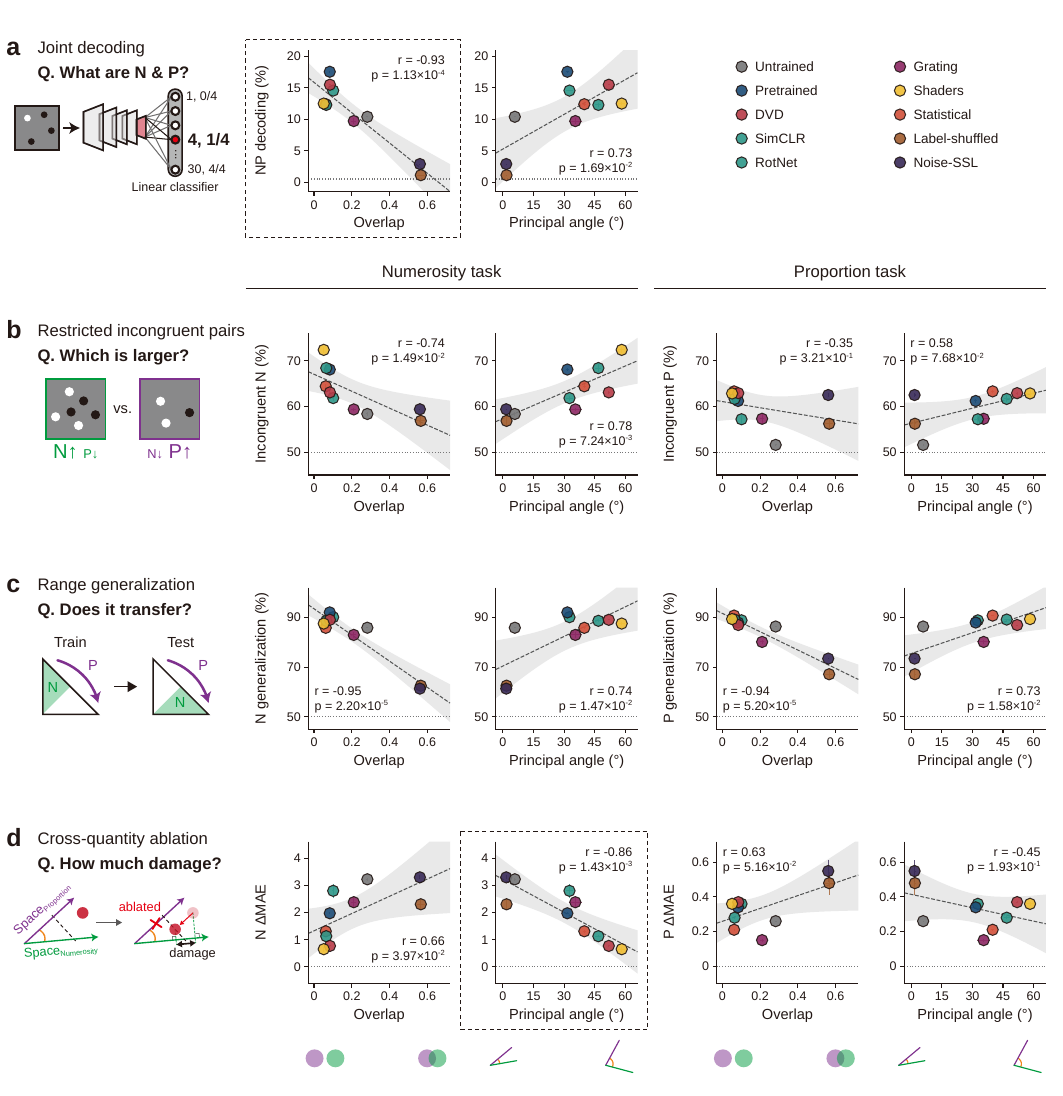}
\caption{Representational interventions across architectures. Activations
of a fixed network were transformed within its joint N/P subspace, and
function was measured on the transformed activations; colors denote the 12
architectures. (a) Principal-angle manipulation: a smooth deformation shifts
the N subspace relative to the P subspace by $-15^{\circ}$ to $+15^{\circ}$
at nearly constant overlap. Insets, AlexNet representation at
$-15^{\circ}$, original, and $+15^{\circ}$ (top, principal-plane view;
bottom, fixed N/P-residual frame). (b) From raw values to the summary curve.
Step~1: decoding accuracy of a retrained readout (top) and cross-quantity
damage (bottom; N, left axis; P, right axis) against the principal angle,
one curve per architecture. Step~2: each curve is centered on its original
representation, with the x-axis expressed as the normalized angle change and
the y-axis as the difference from the original value. Steps~3 and~4: values
are expressed as ratios to the original and averaged over architectures,
giving Figure~\ref{fig:main-4}c; the $-15^{\circ}$ step was infeasible for
DenseNet121. (c) Overlap manipulation: within-condition spread is rescaled
($\times 0.25$ to $\times 4$) around unchanged condition centers, changing
overlap at a fixed principal angle. Insets, AlexNet representation at each
scale. (d) As in (b), against overlap and, after centering, the relative
within-condition RMS, giving Figure~\ref{fig:main-4}d. Solid lines, N;
dashed lines, P; open markers, original representations; asterisks,
rank-matched random-subspace ablation of the original representation.}
\label{fig:supp-9}
\end{figure}

\clearpage
\begin{figure}[!htbp]
\centering
\includegraphics[width=1.00\linewidth]{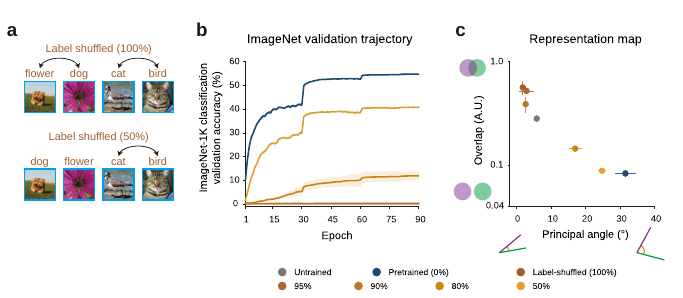}
\caption{Correct-label correspondence links visual learning to quantity
separation. (a) Schematic of complete and partial label reassignment.
(b) ImageNet validation accuracy increases as the shuffled-label
fraction decreases, demonstrating more effective learning from stronger
label correspondence. (c) The conditions span a continuum from fully
Label-shuffled through the vicinity of Untrained to Pretrained in
principal-angle--overlap space. Decreasing the shuffled-label fraction
increases the principal angle and reduces overlap, linking label
correspondence to improved quantity separation. Percentages denote
shuffled-label fractions, from 0\% (Pretrained) to 100\% (fully shuffled).}
\label{fig:supp-10}
\end{figure}

\clearpage
\begin{figure}[!htbp]
\centering
\includegraphics[width=0.54\linewidth]{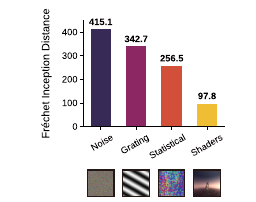}
\caption{Synthetic-image similarity to ImageNet-1K.
Fr\'{e}chet Inception Distance (FID; \citealp{heusel2017gans}) evaluates
the similarity of Noise, Grating, Statistical, and Shaders image distributions
to ImageNet-1K;
lower values indicate greater similarity. Example images are shown below
the bars. Across these visual diets, lower FID generally accompanies
larger principal angles and lower overlap after training (Figure~\ref{fig:main-5}
and Figure~\ref{fig:supp-12}). This pattern supports the main-text account that
learning sufficiently rich visual structure, rather than natural-image
origin alone, promotes quantity separation.}
\label{fig:supp-11}
\end{figure}

\clearpage
\begin{figure}[!htbp]
\centering
\includegraphics[width=0.92\linewidth]{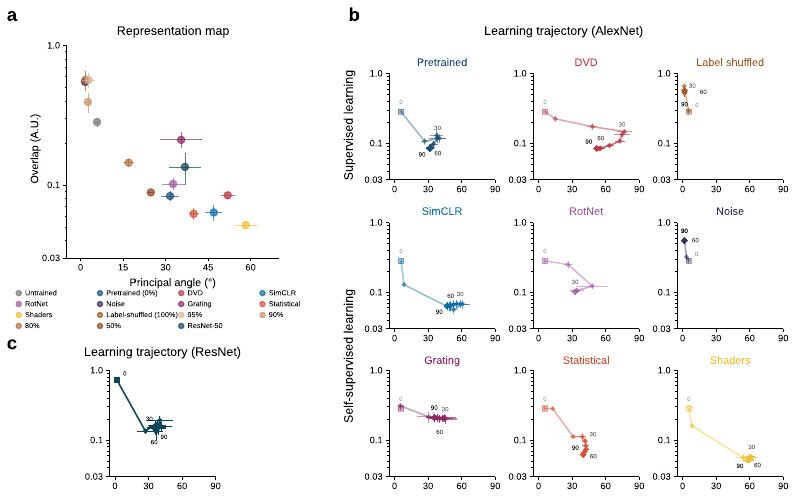}
\caption{Learning trajectories in principal-angle--overlap space.
(a) Summary representation map of the network states and label-correspondence
conditions. (b) State-specific AlexNet trajectories during supervised and
self-supervised learning. (c) The corresponding trajectory for ResNet-50.
Connected points track geometry across training checkpoints; numeric
annotations indicate epochs. Rightward and downward movements indicate
larger principal angles and lower overlap, respectively. The trajectories
show how different visual-learning conditions reshape quantity geometry
relative to their initial representations.}
\label{fig:supp-12}
\end{figure}

\clearpage
\begin{figure}[!htbp]
\centering
\includegraphics[width=1.00\linewidth]{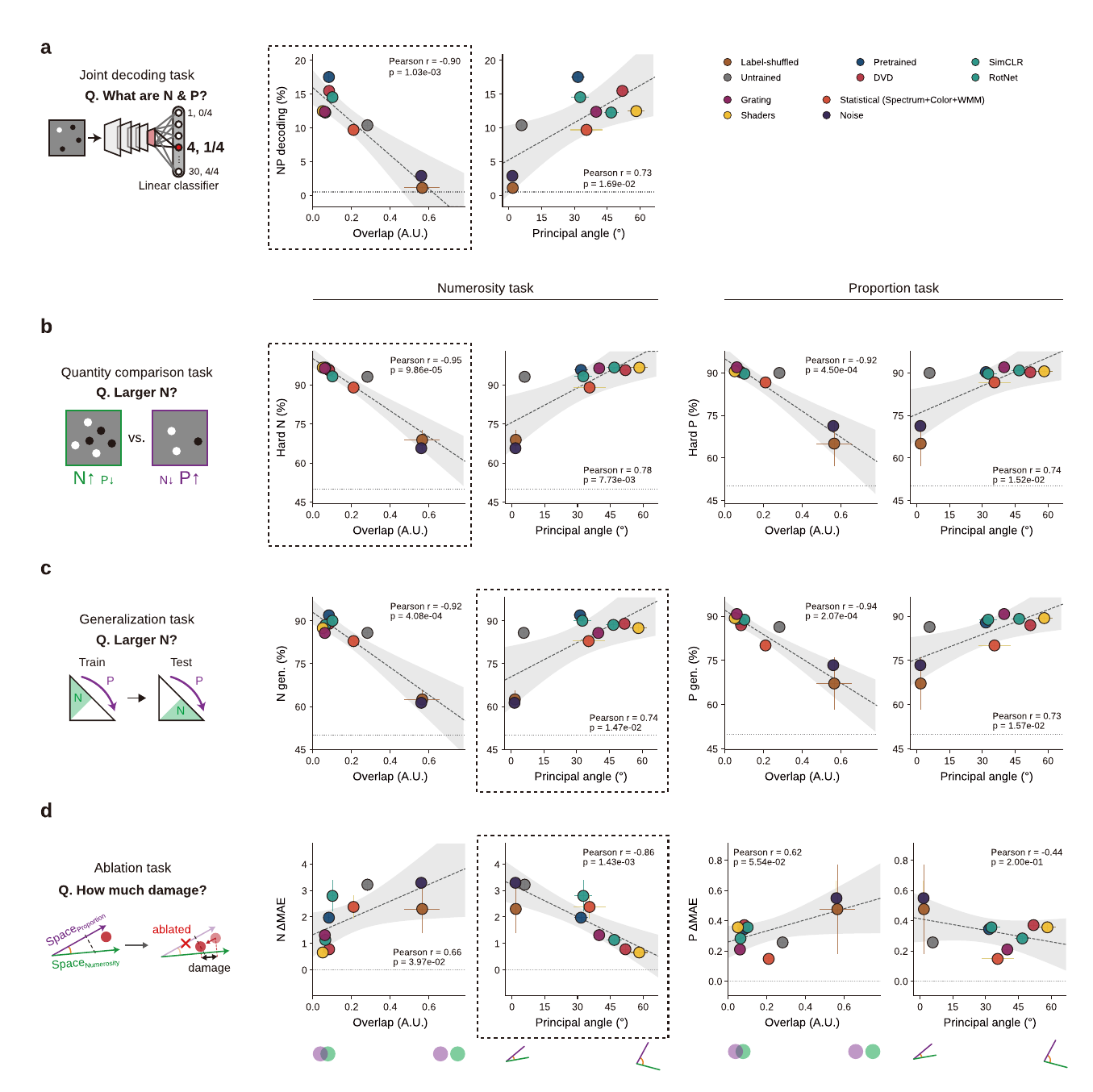}
\caption{Geometry--function correlation matrix. (a) Exact joint NP decoding.
(b) Restricted pairwise N and P comparisons. (c) Cross-context generalization
of N and P comparison rules. (d) Cross-quantity damage after mean-preserving
subspace ablation (Figure~\ref{fig:supp-3}), measured as the change in
ordinal MAE. Each functional readout is shown against condition-wise overlap
and principal angle. Points represent state means; the annotated Pearson
correlations describe cross-state associations. Dashed fits and shaded bands
show ordinary least-squares trends and 95\% confidence intervals. Boxed
panels correspond to the selected relationships in
Figure~\ref{fig:main-7}.}
\label{fig:supp-13}
\end{figure}

\clearpage
\begin{figure}[!htbp]
\centering
\includegraphics[width=1.00\linewidth]{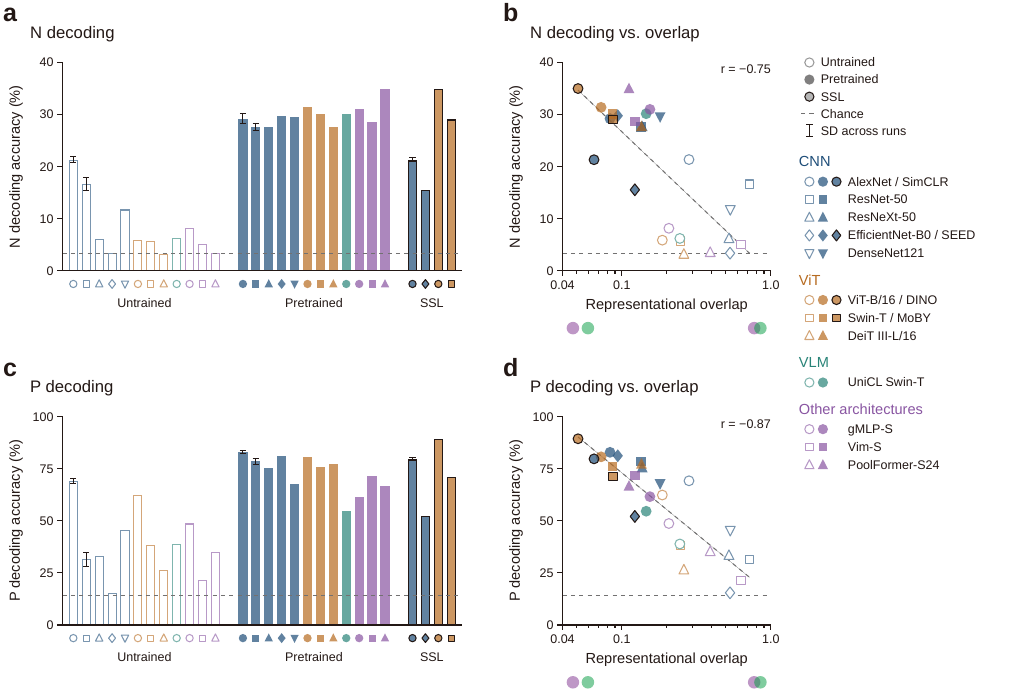}
\caption{N and P decoding across architectures. (a,c) N and P decoding
accuracy for the Untrained, Pretrained, and self-supervised (SSL) states of
12 architectures; marker shapes and colors identify the models. Error bars
show SD across runs where several runs were available (AlexNet and
ResNet-50); dashed lines indicate chance. (b,d) Decoding accuracy against
representational overlap (log scale). Dashed lines show least-squares fits,
and $r$ is the Pearson correlation with log overlap across the 28 states.
These correlations are descriptive, because the states are not independent
architectures.}
\label{fig:supp-14}
\end{figure}

\end{document}